%% file: emnlp2022.tex
\pdfoutput=1
\documentclass[11pt]{article}

\usepackage{EMNLP2022}

\usepackage{times}
\usepackage{latexsym}
\usepackage{graphicx}
\usepackage{caption}
\usepackage{multirow}
\usepackage{multicol}
\usepackage{booktabs}
\usepackage{amsmath}
\usepackage[T1]{fontenc}
\usepackage[utf8]{inputenc}

\usepackage{microtype}

\usepackage{inconsolata}
\usepackage{comment}

\newcommand{\squad}{\textsc{SQuAD 2.0}}
\newcommand{\drop}{\textsc{DROP}}
\newcommand{\multirc}{\textsc{MultiRC}}
\newcommand{\qasc}{\textsc{QASC}}
\newcommand{\mathqa}{\textsc{MathQA}}
\newcommand{\hotpot}{\textsc{HotpotQA}}
\newcommand{\strategy}{\textsc{StrategyQA}}
\newcommand{\svamp}{\textsc{SVAMP}}
\newcommand{\breakqa}{\textsc{Break}}
\newcommand*\samethanks[1][\value{footnote}]{\footnotemark[#1]}

\title{Is a Question Decomposition Unit All We Need?}

\author{First Author \\
  Affiliation / Address line 1 \\
  Affiliation / Address line 2 \\
  Affiliation / Address line 3 \\
  \texttt{email@domain} \\\And
  Second Author \\
  Affiliation / Address line 1 \\
  Affiliation / Address line 2 \\
  Affiliation / Address line 3 \\
  \texttt{email@domain} \\}

\author{Pruthvi Patel\thanks{~~Equal Contribution} \quad Swaroop Mishra\samethanks \quad Mihir Parmar \quad Chitta Baral \vspace{5mm} \\ Arizona State University}

\begin{document}
\maketitle
\begin{abstract}
Large Language Models (LMs) have achieved state-of-the-art performance on many Natural Language Processing (NLP) benchmarks. With the growing number of new benchmarks, we build bigger and more complex LMs. However, building new LMs may not be an ideal option owing to the cost, time and environmental impact associated with it. We explore an alternative route: can we modify data by expressing it in terms of the model's strengths, so that a question becomes easier for models to answer? We investigate if humans can decompose a hard question into a set of simpler questions that are relatively easier for models to solve. We analyze a range of datasets involving various forms of reasoning and find that it is indeed possible to significantly improve model performance (24\% for GPT3 and 29\% for RoBERTa-SQuAD along with a symbolic calculator) via decomposition. Our approach provides a viable option to involve people in NLP research in a meaningful way. Our findings indicate that Human-in-the-loop Question Decomposition (HQD) can potentially provide an alternate path to building large LMs\footnote{https://github.com/Pruthvi98/QuestionDecomposition}.
\end{abstract}

\input{sections/introduction}

\input{sections/related_work}

\input{sections/methods}

%%%%%%%%%%%%% EXPERIMENTS %%%%%%%%%%%%%%%%%%%%%%%
\input{sections/Experiemental_Setup}

\input{sections/results_analysis}

\input{sections/conclusions}

\section*{Limitations}
Our human-in-the-loop methodology shows promising results by decomposing questions, however, certain questions are still difficult to decompose for humans as well. For instance, the question \textit{"Which country is New York in?"}, is hard to decompose further. Determining which questions to decompose is also an important challenge and under-explored in this work. Furthermore, decomposed questions in the chain which have more than one correct answers might lead to an incorrect final answer. Automating the process of decomposition while addressing these issues is a promising area for future work.

\bibliography{anthology,custom}
\bibliographystyle{acl_natbib}

\clearpage

%\subsection{Appendices}
\appendix
\input{sections/appendix_prompts}


\end{document}

%% file: sections/introduction.tex
\section{Introduction}
\label{sec:introduction}

With the advent of large LMs, we have achieved state-of-the-art performance on many NLP benchmarks \citep{gpt2, gpt3, t0pp}. Our benchmarks are evolving and becoming harder over time. To solve new benchmarks, we have been designing more complex and bigger LMs at the cost of computational resources, time and its negative impact on the environment. Building newer LMs for solving new benchmarks may not be an ideal and sustainable option over time. Inspired by humans, who often view new tasks as a combination of existing tasks, we explore if we can mimic humans and help the model solve a new task by decomposing~\cite{mishra2021reframing} it as a combination of tasks that the model excels at and already knows. 

As NLP applications are increasingly more and more popular among people in their daily activities, it is essential to develop methods that involve humans in NLP-powered applications in meaningful ways. Our approach attempts to fill this gap in LMs by providing a human-centric approach to modifying data. 
Solving complex QA tasks such as multi-hop QA, and numerical reasoning has been a challenge for models. Question Decomposition (QD) has recently been explored to empower models to solve these tasks with the added advantage of interpretability. However, previous studies on QD are limited to some specific datasets~\cite{modularqa:journals/corr/abs-2009-00751} such as \drop{}~\cite{DROP:journals/corr/abs-1903-00161} and \hotpot{}~\cite{hotpot:journals/corr/abs-1809-09600}. We analyze a range of datasets involving various forms of reasoning to investigate if ``\textit{a Question Decomposition Unit All We Need?}"

Figure \ref{fig:main_figure} shows the schematic representation of a QD unit. The \textcolor{blue}{original question} is difficult for a model to answer. However, it becomes easier for the model when a human decomposes the question into a set of \textcolor{teal}{simpler questions}.

% Question decomposition in the literature has been limited to 

% Also, new LMs are pre-trained with enormous amount of data compared to previous models. However, building newer LMs for solving new benchmarks is not feasible in real-life. We believe that rather than increasing model size and complexity, we can modify data in a way that LMs understand well. In this paper, we focus on Question Answering (QA) task to analyze our hypothesis.

% The current state of the art Large Language Models in NLP have shown massive performance gains by using bigger models bigger trained on larger amounts of data. Hence, using this as a guiding principle, each newer model is bigger or/and uses more enormous amounts of data than the previous model \citep{gpt2, gpt3, t0pp}. 

% We can observe that a question is broken down into two sub-questions such that each question has a reference in the passage, and the final answer is computed as an operation over the two or more sub-answers. 

\input{figures/main_figure}

We manually decompose randomly selected 50 samples of each dataset. The decompositions we perform are purely based on intuitions to reduce the complexity of the question, inspired by the success of task-level instruction decomposition~\cite{mishra2021reframing} in improving model performance. We experiment with GPT3~\cite{gpt3} and RoBERTA~\cite{liu2019roberta} fine-tuned on SQuAD 2.0~\citep{rajpurkar2018know} and find that HQD significantly improves model performance (24\% for GPT-3 and 29\% for RoBERTa-SQuAD along with a symbolic calculator). Here, the evaluation happens on unseen tasks on which the model is not fine-tuned. Our findings indicate that Human-in-the-loop Question Decomposition (HQD) can potentially provide an alternate path to building large LMs. We hope our work will encourage the community to develop human-centric solutions that actively involve humans while leveraging NLP resources.

%% file: figures/main_figure.tex
\begin{figure*}[ht]
    \centering
    \includegraphics[width=0.7\linewidth]{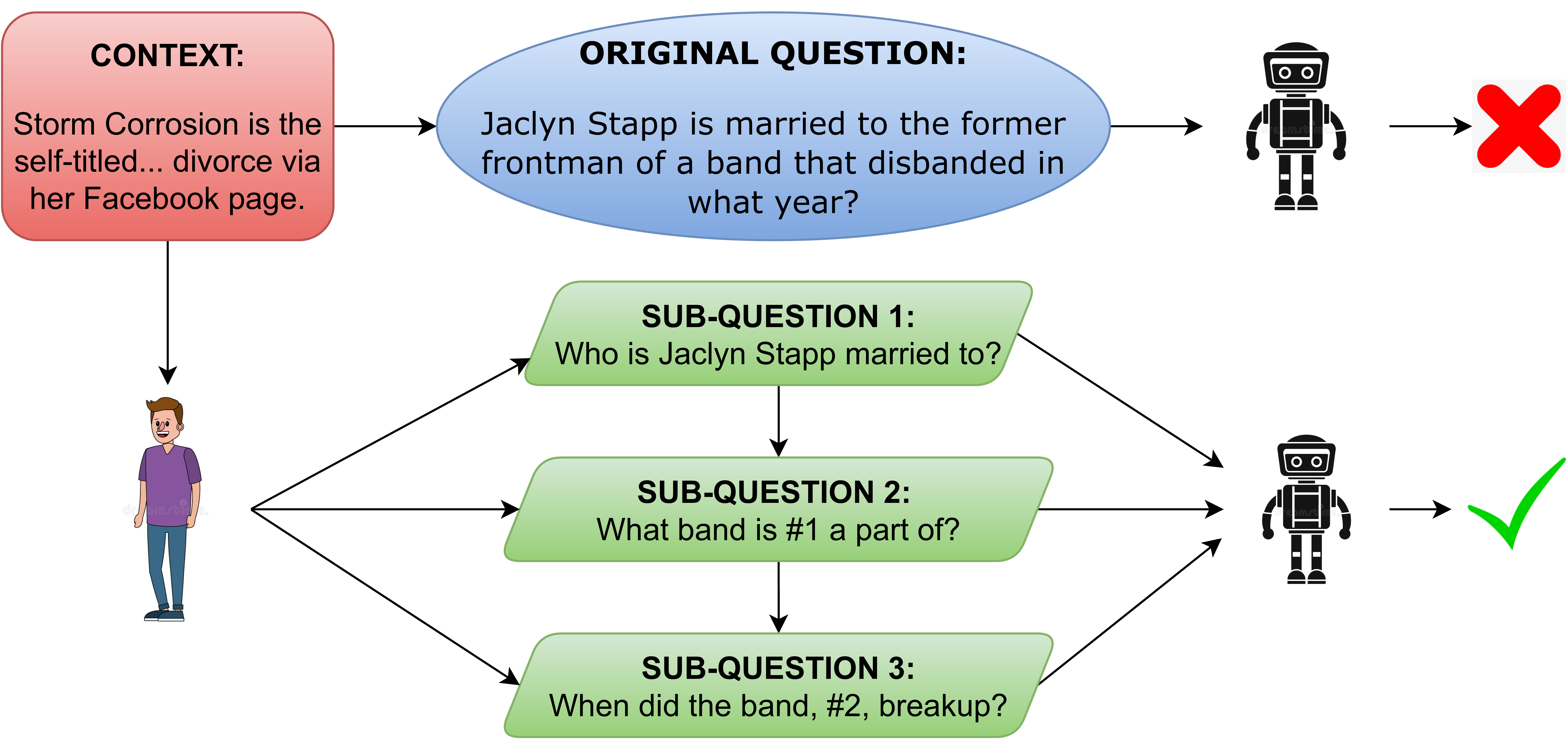}
    \caption{The \textcolor{blue}{original question} is answered incorrectly by a model. A human then decomposes the question into a set of \textcolor{teal}{simpler questions} which the model then answers correctly.}
    \label{fig:main_figure}
\end{figure*}

%% file: sections/related_work.tex
\section{Related Work}
\label{sec:related_work}

% \paragraph{Question Answering} The complexity in reasoning involved in question answering datasets has been increasing ever since \citep{rajpurkar2018know} introduced SQuAD. SQuAD dataset has questions which can be answered using a single sentence in the given paragraph, which is known as single-hop reading comprehension. Recently, datasets like HotpotQA \cite{hotpot:journals/corr/abs-1809-09600}, DROP \citep{DROP:journals/corr/abs-1903-00161} and MultiRC \citep{MultiRC2018} have introduced questions which require both implicit and explicit reasoning over multiple sentences. Furthermore, complexity is introduced by designing questions which need mathematical \citep{mathqa, svamp}, strategic \citep{strategyqa} and fact-based reasoning over multiple choices \citep{qasc}.

% Different from using LMs for reading comprehension question-answer pair~\citep{gpt2, gpt3}, 
% Question decomposition aims to decompose a complex question into multiple single-hop questions~\citep{complexwebqa, decomprc}.
% A recent methodology to reason over multiple sentences in reading comprehension datasets is to decompose the question into single-hop questions \citep{complexwebqa, decomprc}.
A recent methodology to reason over multiple sentences in reading comprehension datasets is to decompose the question into single-hop questions \citep{complexwebqa, decomprc}. 
\citet{decomprc} decompose questions from \hotpot{} using span predictions based on reasoning types and picks the best decomposition using a decomposition scorer. \citet{modularqa:journals/corr/abs-2009-00751} generate decompositions by training a BART model on question generation task by providing context, answers and hints. \citet{break} crowd-sourced annotations for decompositions of questions. \citet{perez2020unsupervised}, on the other hand, uses the unsupervised mechanism of generating decomposition by mapping a hard question to a set of candidate sub-questions from a question corpus. \citet{iyyer-search-based} answer a question sequentially using a neural semantic parsing framework over crowdsourced decompositions for questions from WikiTableQuestions. Decomposition using text-to-SQL query conversion has also been studied \citep{text2sql}. Also, knowledge graphs are combined with neural networks to generate decompositions \citep{denotational_semantics_gupta}. Recently, \citet{decompositional_probes} presented another use case where decompositions can be used to probe models to create explanations for their reasoning. 

% All these methodologies focus on mainly HotpotQA and DROP datasets involing multi-hop reasoning in reading comprehensions. We expand the types of reasoning that can benefit from the success obtained using decomposition.

%% file: sections/methods.tex
\section{Methods}
\label{sec: methods}

\subsection{Datasets}

We select eight datasets covering a diverse set of reasoning skills and domains: (1) \hotpot{} \citep{hotpot:journals/corr/abs-1809-09600}, (2) \drop{} \citep{DROP:journals/corr/abs-1903-00161}, (3) \multirc{} \citep{MultiRC2018}, (4) \strategy{} \citep{strategyqa}, (5) \qasc{} \cite{qasc}, (6) \mathqa{} \cite{mathqa}, (7) \svamp{} \cite{svamp}, and (8) \breakqa{} \cite{break}. Table \ref{tab:datasets_type} indicates the different task types for each dataset. 

\input{tables/dataset_info}

\subsection{Decomposition Process}

For each dataset, we randomly select 50 instances for manual decomposition. The question in each dataset is decomposed into two or more questions. Table~\ref{tab:decomp1},~\ref{tab:decomp2}, ~\ref{tab:decomp3} and ~\ref{tab:decomp4} show examples of decomposition for various datasets. For each dataset, we created a set $\mathcal{D}$ for decomposed questions. Each element $\mathcal{D}_i \in \mathcal{D}$ can be represented as below:

\begin{equation*}
    \mathcal{D}_i = \{\mathcal{C}_i, \mathcal{Q}_i, \mathcal{Q}_d, \mathcal{A}_i, \mathcal{A}_d\},
\end{equation*}
 
where $\mathcal{C}_i$ is the context paragraphs, $\mathcal{Q}_i$ is the original question, $\mathcal{Q}_d$ is the set of decomposed questions, $\mathcal{A}_i$ is an original answer, and $\mathcal{A}_d$ is the set of answers for corresponding decomposed questions. For questions that require arithmetic or logical operations, we use a computational unit as suggested in \citet{modularqa:journals/corr/abs-2009-00751}, which takes a decomposed question as input in the following format:

\begin{equation*}
    \{\mathcal{O}\} ! \#m_1 ! \#m_2 ! .... ! \#m_n,
\end{equation*}

where $\mathcal{O} = $ \{summation, difference, division, multiplication, greater, lesser, power, concat, return, remainder\}, \#$m_i$ are answers of previous decomposed questions and $!$ separates the operands.

% After the first decomposed question \(Q_{i0}\), each question might contain the answer extracted from the context of previous decomposed questions. The portion of the question to be replaced by the previous answer/s is denoted using \# followed by the question number whose answer needs to be replaced.

% Hence, each subsequent decomposed question is a function of \(Q_{i}\), \(Q_{ij}\) and \(A_{ik}\), such that \(k < j\).

%% file: tables/dataset_info.tex
\begin{table}[t]

\centering 
\setlength\tabcolsep{4.0pt}
\footnotesize

\resizebox{\linewidth}{!}{

    \begin{tabular}{ c | c }
      \toprule
      \textbf{Name} & \textbf{Type}\\
      \midrule
      \hotpot & Multihop RC\\
      %\hline
      \drop & Mulithop RC\\
      %\hline
      \strategy & Strategic Reasoning\\
      %\hline
      \multirc & RC\\
      %\hline
      \breakqa & RC\\
      %\hline
      \mathqa & Mathematical Reasoning\\
      %\hline
      \qasc & Fact-based Multichoice\\
      %\hline
      \svamp & Context-based Math Word Problems\\
      \bottomrule
    \end{tabular}
}
\caption{Type of QA task corresponding to each dataset. RC: Reading Comprehension}
\label{tab:datasets_type}
    
\end{table}

%% file: sections/Experiemental_Setup.tex
\section{Experimental Setup}
\label{sec:experimental_setup}

\paragraph{Models} We use GPT-3 \cite{gpt3} to generate answers for original and decomposed questions. To show that QD significantly improves performance even on simpler models, we use RoBERTa-base finetuned on \squad{} dataset (i.e., RoBERTa-SQuAD). Additionally, we use RoBERTa-base finetuned on BoolQ dataset \cite{clark2019boolq} (i.e., RoBERTa-BoolQ) for original and decomposed questions in \strategy{} since they are True/False type questions.

\paragraph{Experiments} To create baselines, we evaluate all models on the original question along with the context. We evaluate all models on the manually decomposed questions in the proposed method. We carry out all experiments in GPT-3 by designing prompts for each dataset\footnote{See Appendix \ref{app:prompts} for more details}. For RoBERTa-based models, we use RoBERTa-SQuAD for \multirc, \breakqa, \hotpot{} and \drop{} datasets, since \squad{} is designed for a reading comprehension task. For \strategy{}, we use two RoBERTa-base models: (1) RoBERTa-BoolQ, which is used to answer the final boolean type of questions, and (2) RoBERTa-SQuAD which is used to answer the remaining decomposition questions. For \svamp, we use the RoBERTa-SQuAD model to extract the necessary operands using decomposed questions and then we use the computational module to perform various operations. In all experiments, we use decomposition to get to the final answer sequentially. 
% when baseline model generates a wrong output. 

%An example is provided in figure 2. For each dataset, we empirically set the temperature as 0.7, top P as 1 and maximum length of 128 with both frequency and presence penalty set to 0.

% \textbf{GPT3 Baseline} For each dataset, we consider as baseline the results obtained from answering only the original question along with the context. We use GPT3 provide both the context and the question in the prompt to generate the required answer.\\\\
% \textbf{Decompose\_GPT3} This setting is used if the answer generated using GPT3 Baseline is incorrect. We use the decomposition available for the question and provide both the context and decomposed question to GPT3.\\

% We further use a RoBERTa model pretrained on SQuAD dataset to compare the results. We do this in order to prove that even without GPT3, one can still achieve significant improvements using a much simpler model. Since SQuAD is a reading comprehension dataset, we use this model on MultiRC, BREAK, HotpotQA and DROP datasets. This baseline is created in the same way as GPT3 Baseline.\\\\
% \textbf{RoBERTa Baseline} This baseline is created in a similar fashion as GPT3 Baseline. We use the deepset/roberta-base-squad2 model to provide a context and the original question to generate an answer.\\\\
% \textbf{Decompose\_RoBERTa} If the RoBERTa baseline generates a wrong output, we use the decompositions to sequentially get to the final answer.\\

\paragraph{Metrics} For all our experiments, we use Rouge-L \cite{lin2004rouge}, $F_1$-score and Exact Match (EM) as the evaluation metrics.

%% file: sections/results_analysis.tex
\section{Results and Analysis}

Here, we divide our datasets into four categories: (1) RC: \hotpot, \drop, \multirc, and \breakqa{} in Reading Comprehension (RC), (2) MATH: \mathqa{} and \svamp{} in Mathematical reasoning , (3) MC: \qasc{} in Multi-Choice QA (MC) , and (4) SR: \strategy{} in Strategy Reasoning (SR). All results presented in this sections are averaged over tasks for each category.

\subsection{Experimental Results}

\paragraph{GPT-3} Figure \ref{fig:gpt_3_results} shows the GPT-3 performance in terms of average $F_1$-scores for each category. From the Figure \ref{fig:gpt_3_results}, we can observe that our proposed approach outperforms baseline by $\sim24\%$. Appendix \ref{app:results} presents all results in terms of $F_1$-scores, EM and Rouge-L for all datasets and categories. 

\input{figures/roberta}

\input{figures/gpt3}

\paragraph{RoBERTa} Figure \ref{fig:fig_roberta} represents the results we obtain using RoBERTa-based models in terms of $F_1$-scores for each category. On an average, we achieve $\sim29\%$ of significant improvement compared to the baseline. Appendix \ref{app:results} presents all results in terms of $F_1$-scores, EM and Rouge-L for all datasets and categories.

\input{sections/analysis}

%% file: figures/roberta.tex
\begin{figure}[ht]
  \centering
  \includegraphics[width=\linewidth]{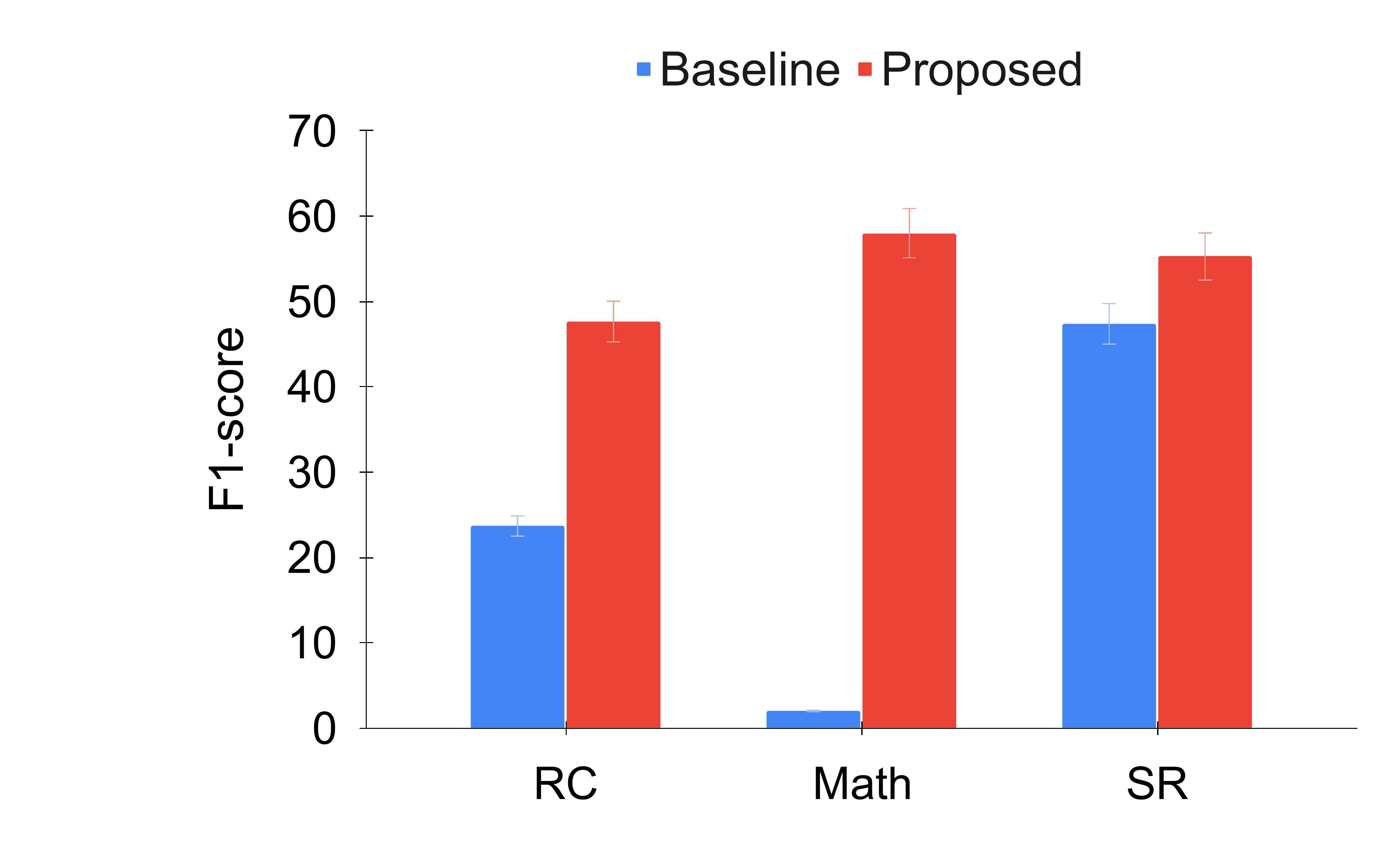}
  \caption{Results in terms of $F_1$-score across different categories for RoBERTa-based models. RC: Reading Comprehension, MATH: Mathematical reasoning, SR: Strategy Reasoning.}
\label{fig:fig_roberta}
\end{figure}

%% file: figures/gpt3.tex
\begin{figure}[ht]
  \centering
  \includegraphics[width=\linewidth]{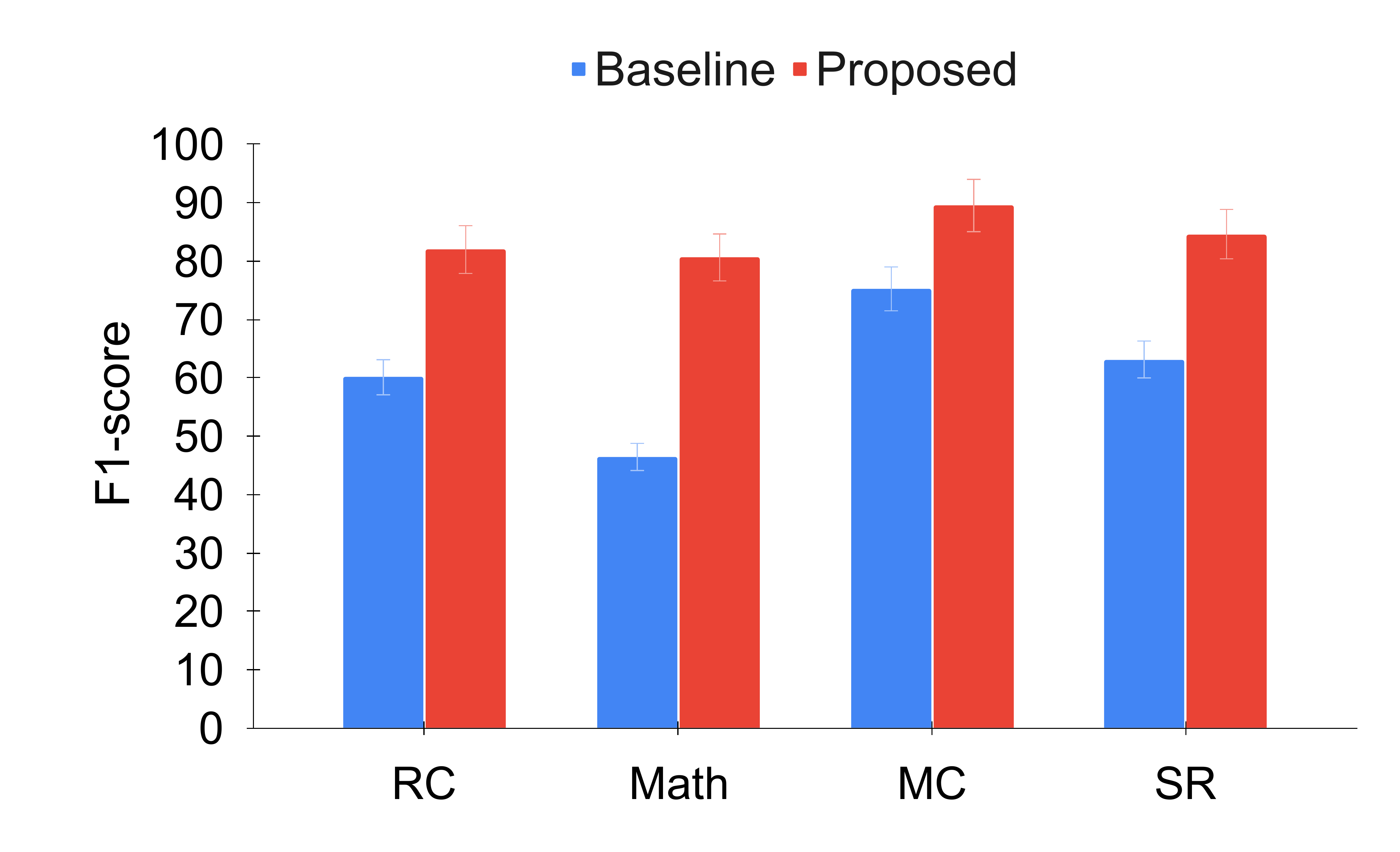}
\caption{Results in terms of $F_1$-score across different categories for GPT-3. RC: Reading Comprehension, MATH: Mathematical reasoning, MC: Multi-Choice QA, SR: Strategy Reasoning.}
\label{fig:gpt_3_results}
\end{figure}

%% file: sections/analysis.tex
\subsection{Analysis}

% \paragraph{Generalization of a Question Decomposition Unit} Figure \ref{fig:gpt_3_results} and Figure \ref{fig:fig_roberta} shows that QD improves performance significantly on variety of QA tasks using GPT-3 and RoBERTa-based models. Such improvement across different kind of tasks indicate the benefit of utilizing decompositions and show its ability to generalize across various tasks.

\paragraph{Customized Question Decomposition for Each Model} There can be multiple ways to decompose a question based on the context. Multiple factors go into deciding how to break down a question. One factor is the strength of the model. For instance, if we use a model finetuned on SQuAD, it might be beneficial to ensure that the decompositions are more granular and are generated to answer from a context span. On the other hand, if we have a more sophisticated model like GPT3, we might not necessarily need to do so. The results shown in Figure~\ref{fig:fig_roberta} are obtained on RoBERTa finetuned on SQuAD by using decompositions originally designed for GPT3; note that in this case, the answers to the decompositions might not always be the span of a particular sentence in the context. However, we achieve a decent performance improvement. We believe the performance gain will be greater if decompositions are designed to match the model's strengths. Examples of such decompositions are included in the Appendix \ref{app:prompts}.

% \paragraph{Leveraging reading comprehension models} \autoref{fig:fig_roberta} shows results obtained by leveraging the RoBERTa model. The results for SVAMP and StrategyQA indicate that decompositions enable us to leverage single-hop reading comprehension models to solve more complex tasks without adding the complexity of models specific to these tasks. Hence, using decomposition, only one simple model can enable us to use a variety of datasets that previously needed complex architectures to be trained and used. SVAMP, for instance, was trained on three different models, each of which predicted both the operands and the equation. Using decompositions would thus enable us to decouple the task of extracting the operands from predicting the operation and make that task more efficient.
\begin{figure}
    \centering
    \includegraphics[width=0.9\linewidth]{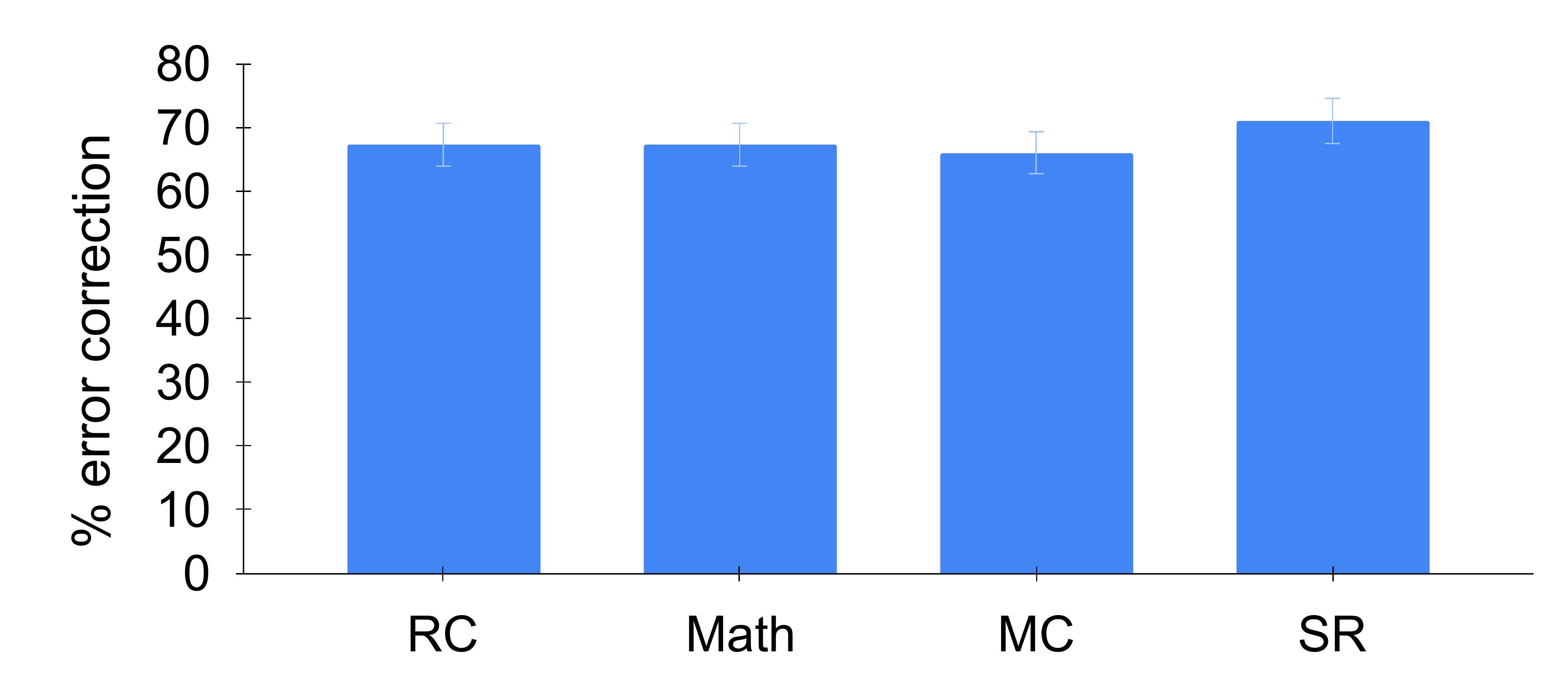}
    \caption{\% error correction by using decompositions with GPT3}
    \label{fig:GPT3_error_correction}
\end{figure}

\begin{figure}
    \centering
    \includegraphics[width=0.7\linewidth]{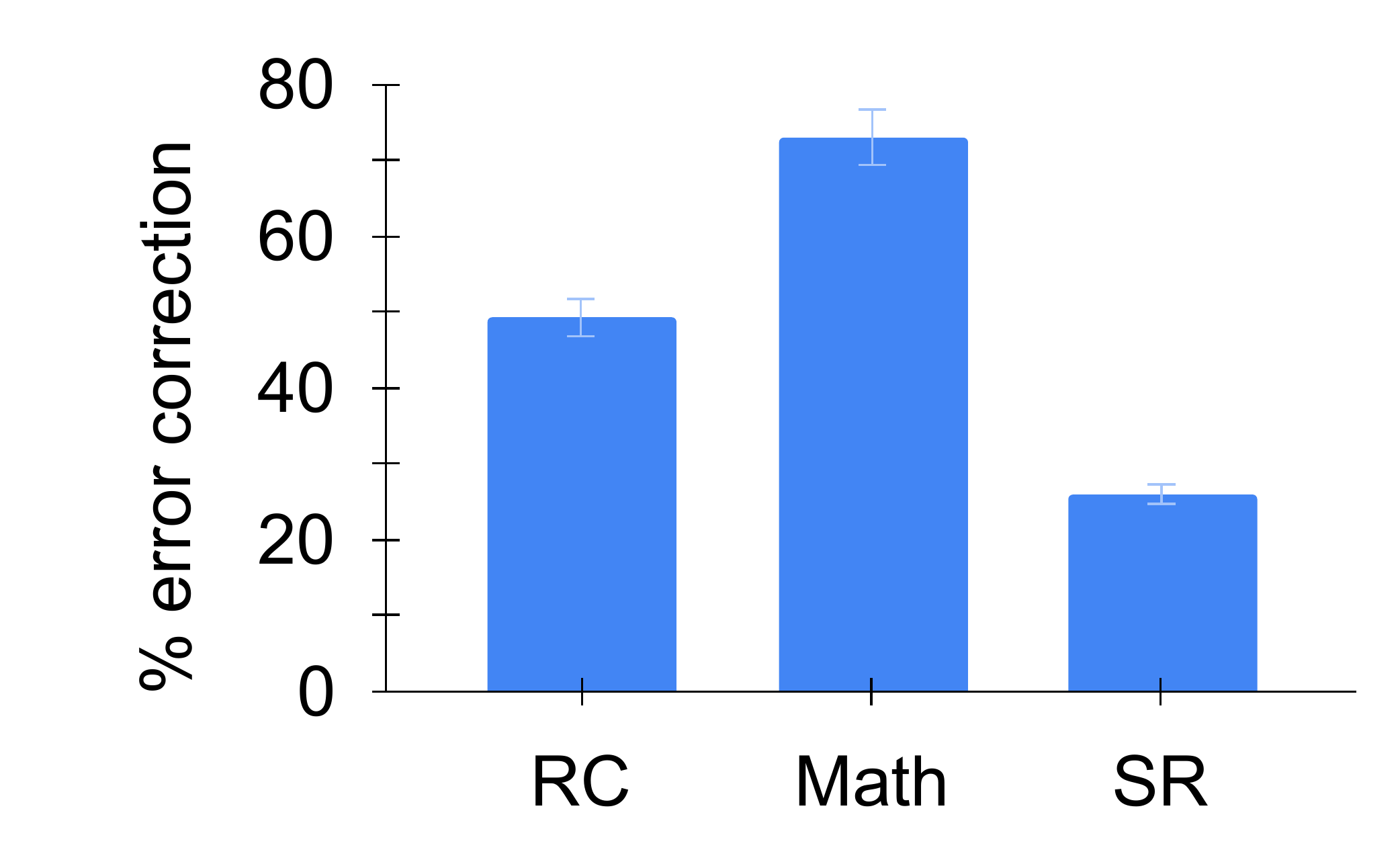}
    \caption{\% error correction by using decompositions with RoBERTa}
    \label{fig:roberta_error_correction}
\end{figure}
\paragraph{Qualitative Analysis} We conduct qualitative analysis to capture the evaluation aspects missed in the automated evaluation metrics. Here, we manually inspect and consider a generated answer to be correct if it is semantically similar to the gold annotation.
Figure \ref{fig:GPT3_error_correction} and \ref{fig:roberta_error_correction} show the contribution of QD in correcting model prediction.
We observe that the decompositions correct more than 60\% of the errors made on the original questions.
% that for all the datasets the percentage of questions that are answered correctly using decompositions that were answered incorrectly using the original question for GPT3, and \autoref{fig:roberta_error_correction} for RoBERTa. As shown, 
% if both the generated answer and the true answer mean the same thing and if the generated answer has the correct answer and additional relevant information not present in the true answer. 
\begin{figure}[ht]
  \centering
  \includegraphics[width=\linewidth]{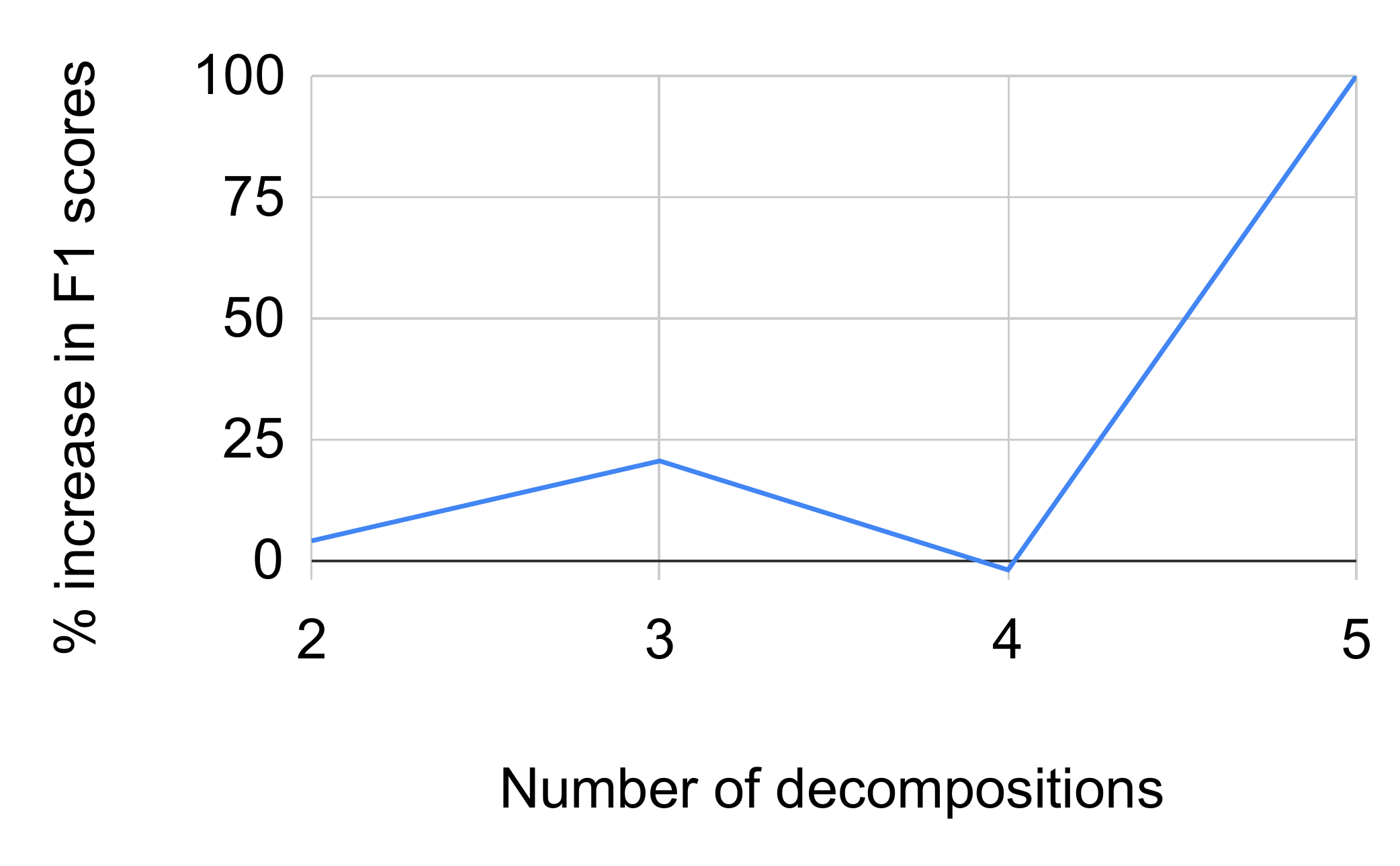}
  \caption{Performance improvements in F1 scores for questions with 2, 3, 4 and 5 decompositions.}
  \label{fig:num_qa}
\end{figure}
\paragraph{Error Analysis} We conduct error analysis and observe that the major source of error is the error propagated from one of the decomposed questions. Errors, in general, are of two types: (i) incorrect span selection and (ii) failure to collect all possible answers in the initial step of decomposition; this often omits the actual correct answer leaving no room for later decomposition units to generate the correct answer. Errors occur in \qasc{} because our method of context-independent decomposition (via intuition) sometimes leads to open-ended questions which models find hard to answer.
% We get incorrect answers in QASC even while using decompositions because without the context, the questions can sometimes be open-ended. Even so, decompositions give the correct answer 66\% of the time. 
Examples of errors have been included in the Appendix \ref{app:error_examples}.

\paragraph{Effect of Decomposition on Math Datasets} 
% \textcolor{blue}{Want to change title of this paragraph}

We observe that Math datasets benefit the most from decomposition. This may be because of two reasons: 1) majority of math questions can be decomposed as a combination of extractive QA (where the answer is a span) and a symbolic calculation. Both of these are strengths of language models (note that we use calculators that provide accurate answers consistently). However, this is not necessarily true in case of other QA tasks. In a decomposition chain, if the answer in one step goes wrong, it propagates till the end and the final prediction becomes wrong. 2) language models by default struggle to do math tasks~\cite{svamp, mishra2022numglue}, so the performance improvement seems more prominent there.

\paragraph{Effect of Number of Decompositions on Results}
We typically decompose a question based on the number of operations associated with it (e.g. mathematical calculation or single hop operation). Increase in the number of decompositions has the advantage that it simplifies the original question, but it can also have the disadvantage that if the answer to one of the questions in the chain is incorrect, the end answer becomes incorrect. This is also evident from our empirical analysis on HOTPOTQA and SVAMP datasets where we observe that there is no direct correlation between the number of labeling QA and the final performance. Figure \ref{fig:num_qa} shows the variation in model performance improvement observed for questions with 2, 3, 4 and 5 decompositions.

\paragraph{Efforts to Automate Decomposition} For \hotpot, \drop, and \svamp, we attempt to automate the decomposition process using GPT3. A limitation for generating decompositions for \hotpot{} is that the context length makes it difficult to provide sufficient examples in prompt. With DROP and \svamp, we observe that GPT-3 often generates incorrect arithmetic operations for the last sub-question. It also often fails to develop coherent decompositions of the questions. We also finetune a BART-base~\cite{lewis2020bart} model on our handwritten decompositions. However, the model overfits and fails to produce meaningful decompositions, probably due to the limited number of training samples (see Appendix \ref{app:automated_examples} for examples, details and results).

%% file: sections/conclusions.tex
\section{Conclusion}
The recent trend of building large LMs may not be sustainable to solve evolving benchmarks. We believe that modifying data samples can significantly help the model improve performance. We study the effect of Question Decomposition (QD) on a diverse set of tasks. We decompose questions manually and significantly improve model performance (24\% for GPT3 and 29\% for RoBERTa-SQuAD along with a symbolic calculator). Our findings indicate that Human-in-the-loop Question Decomposition (HQD) can potentially provide an alternate path to building large LMs. 
Our approach provides a viable option to involve people in NLP research. 
We hope our work will encourage the community to develop human-centric solutions that actively involve humans while leveraging NLP resources.

% We plan to apply question decomposition approach to other complex tasks such as logical reasoning~\cite{banerjee2020can,clark2021transformers}, other domains such as biomedical~\cite{parmar2022boxbart}, and explore the potential application of improving model's generalization~\cite{gokhale2022generalized}. 

%% file: sections/appendix_prompts.tex
\section{Prompts}
\label{app:prompts}
Due to the success of large LMs, prompt-based learning is becoming popular to achieve generalization and eliminate the need of creating task-specific models and large scale datasets \cite{liu2021pre}. Recently, instructional prompts have been pivotal in improving the performance of LMs and achieving zero-shot generalization \cite{mishra2021cross, wei2021finetuned, sanh2021multitask, wei2022chain, ouyang2022training, parmar2022boxbart, puri2022many, kuznia2022less, luo2022biotabqa}. We present the instructional prompts that we used to generate answers for various datasets.
\input{sections/drop_hotpot_break_prompt}
\input{sections/mathqa_prompt}
\input{sections/svamp_prompt}
\input{sections/strategyqa_prompt}
\input{sections/qasc_prompt}
\input{sections/multirc_prompt}
\input{sections/error_examples}

\input{sections/automated_examples}

\input{tables/automated_examples}
\input{sections/results}

%% file: sections/drop_hotpot_break_prompt.tex
\subsection{\hotpot, \drop, \breakqa}

Given a context, answer the question using information and facts present in the context. Keep the answer short.\\
Example:\\
Input:\\
Mehmed built a fleet to besiege the city from the sea .Contemporary estimates of the strength of the Ottoman fleet span between about 110 ships , 145 , 160 , 200-250  to 430 . A more realistic modern estimate predicts a fleet strength of 126 ships comprising 6 large galleys, 10 ordinary galleys, 15 smaller galleys, 75 large rowing boats, and 20 horse-transports.:44 Before the siege of Constantinople, it was known that the Ottomans had the ability to cast medium-sized cannons, but the range of some pieces they were able to field far surpassed the defenders' expectations. Instrumental to this Ottoman advancement in arms production was a somewhat mysterious figure by the name of Orban , a Hungarian .:374 One cannon designed by Orban was named "Basilica" and was 27 feet  long, and able to hurl a 600lb  stone ball over a mile .\\
Question: How many ordinary galleys and large rowing boats is estimated from the fleet strength?
Output:\\
Answer: 85\\\\
Input:\\
Context: <<CONTEXT>>\\
Question: <<QUESTION>>\\
Output:\\
Answer: <<OUTPUT GENERATED BY GPT3>>
\input{tables/hotpot_drop_break}

%% file: tables/hotpot_drop_break.tex
\begin{table*}[htp]
    \centering 
    \setlength\tabcolsep{4.0pt}
    \footnotesize
\resizebox{0.8\linewidth}{!}{
    \begin{tabular}{c p{30em}}
    \toprule
    \multirow{10}{*}{HotpotQA} & \textbf{\emph{Context}}: The Larkspur Press is a small letter-press publisher based in Monterey, Kentucky .... , The film also features appearances by Helen Keller, Anne Sullivan, Kate Adams Keller and Phillips Brooks Keller as themselves.  The movie was directed by George Foster Platt and written by Francis Trevelyan Miller.\\
    & \textbf{\emph{Original Question}}: Are John O'Hara and Rabindranath Tagore the same nationality?\\
    & \textbf{\emph{True Answer}}: no\\
    & \textbf{\emph{Decomposed Question 1}}: What is John O'Hara's nationality?\\
    & \textbf{\emph{Generated Answer}}: American\\
    & \textbf{\emph{Decomposed Question 2}}: What is Rbindranath Tagore's nationality?\\
    & \textbf{\emph{Generated Answer}}: Indian\\
    & \textbf{\emph{Decomposed Question 3}}: Is \#1 and \#2 the same nationality?\\
    & \textbf{\emph{Generated Answer}}: No\\
    \hline
    \multirow{10}{*}{DROP} & \textbf{\emph{Context}}: Mehmed built a fleet to besiege the city from the sea .... and able to hurl a 600 lb  stone ball over a mile .\\
    & \textbf{\emph{Original Question}}: How many ordinary galleys and large rowing boats is estimated from the fleet strength?\\
    & \textbf{\emph{True Answer}}: 85\\
    & \textbf{\emph{Decomposed Question 1}}: How many ordinary galleys were there?\\
    & \textbf{\emph{Generated Answer}}: 10\\
    & \textbf{\emph{Decomposed Question 2}}: How many large rowing boats were there?\\
    & \textbf{\emph{Generated Answer}}: 75\\
    & \textbf{\emph{Decomposed Question 3}}: summation ! \#1 ! \#2\\
    & \textbf{\emph{Generated Answer}}: 85\\
    \bottomrule
    \end{tabular}
    }
    \caption{Examples for DROP and HotpotQA.}
    \label{tab:decomp1}
\end{table*}

%% file: sections/mathqa_prompt.tex
\subsection{MATHQA}
Prompt for the original question:\\
Given a problem and 5 options, return the correct option. In order to choose the correct option, you will have to perform some mathematical operations based on the information present in the problem. Look at the examples given below to understand how to answer.\\\\
Input:
Problem: the volume of water inside a swimming pool doubles every hour . if the pool is filled to its full capacity within 8 hours , in how many hours was it filled to one quarter of its capacity\\
Options: a ) 2, b ) 4, c ) 5, d ) 6, e ) 7\\
Output:\\
Answer: 6\\\\
Input:\\
Problem: a train 200 m long can cross an electric pole in 5 sec and then find the speed of the train ?\\
Options: a ) 114 , b ) 124 , c ) 134 , d ) 144 , e ) 154\\
Output:\\
Answer: 144\\\\
Input:\\
Problem: <<Problem>>\\
Options: <<options>>\\
Output:\\
Answer: <<OUTPUT GENERATED BY GPT3>>

%% file: sections/svamp_prompt.tex
\subsection{SVAMP}
Prompt used for both decomposed questions and original questions. The examples contain both decomposed type questions and original type questions.\\\\
Given some context, answer a given question. Use the examples given below as reference.\\\\
Example 1:\\
Input:\\
Context: It takes 4.0 apples to make 1.0 pie.\\
Question: How many apples does it take to make 504.0 pies?\\
Output:\\
Answer: 2016\\\\
Example 2:\\
Input:\\
Context: Mary is baking a cake.The recipe calls for 7.0 cups of flour and 3.0 cups of sugar.She already put in 2.0 cups of flour.\\
Question: How many cups of flour did recipe called?\\
Output:\\
Answer: 7\\\\
Example 3:\\
Input:\\
Context: Each pack of dvds costs 76 dollars. If there is a discount of 25 dollars on each pack
Question: How much is each pack of dvds without the discount?\\
Output:\\
Answer: 76\\\\
Example 4:\\
Input:\\
Context: Conner has 25000.0 dollars in his bank account.Every month he spends 1500.0 dollars.He does not add money to the account.\\
Question: How many dollars Conner spends every month?\\
Output:\\
Answer: 1500\\\\
Input:\\
Context: <CONTEXT>>\\
Question: <<QUESTION>>\\
Output:\\
Answer: <<OUTPUT GENERATED BY GPT3>>
\input{tables/svamp}

%% file: tables/svamp.tex
\begin{table*}[htp]
    \centering 
    \setlength\tabcolsep{4.0pt}
    \footnotesize

\resizebox{0.8\linewidth}{!}{
    \begin{tabular}{c p{30em}}
    \toprule
    \multirow{9}{*}{SVAMP}& \textbf{\emph{Context}}: Bryan took a look at his books as well.If Bryan has 56.0 books in each of his 9.0 bookshelves.\\
    & \textbf{\emph{Original Question}}: How many books does he have in total?\\
    & \textbf{\emph{Answer}}: 504.0\\
    & \textbf{\emph{Decomposed Question 1}}:How many books in each bookshelf?\\
    & \textbf{\emph{Answer}}: 56.0\\
    & \textbf{\emph{Decomposed Question 2}}:How many bookshelves?\\
    & \textbf{\emph{Answer}}: 9.0\\
    & \textbf{\emph{Decomposed Question 3}}: multiplication ! \#1 ! \#2\\
    & \textbf{\emph{Answer}}: 504.0\\
    \hline
    \multirow{9}{*}{MATHQA}& \textbf{\emph{Problem}}: if a train , travelling at a speed of 180 kmph , crosses a pole in 6 sec , then the length of train is ?\\
    & \textbf{\emph{Options}}: a ) 300 , b ) 125 , c ) 288 , d ) 266 , e ) 121\\
    & \textbf{\emph{Annotated Formula}}: multiply(multiply(180, const\_0.2778), 6)\\
    & \textbf{\emph{Answer}}: 300\\
    & \textbf{\emph{Generated Answer}}: 266\\
    & \textbf{\emph{Decomposed Question 1}}: multiplication ! 0.2778 ! 180\\
    & \textbf{\emph{Answer}}: 50.004\\
    & \textbf{\emph{Decomposed Question 2}}: multiplication ! 50.004 ! 6\\
    & \textbf{\emph{Answer}}: 300\\
    \bottomrule
    \end{tabular}
    }
    \caption{Decomposition Examples for SVAMP and MathQA. We use the annotated formula presented in the dataset to make our decompositions.}
    \label{tab:decomp2}
\end{table*}

%% file: sections/strategyqa_prompt.tex
\subsection{StrategyQA}

Input:\\
Context: A melodrama is a dramatic work ...The passengers' response to the hijacking has come to be invested with great moral significance.\\
Question: What do tearjerkers refer to?\\
Output:\\
Answer: a story, song, play, film, or broadcast that moves or is intended to move its audience to tears.\\\\
Input:\\
Context: The purpose of the course is learning to soldier as ... The main motor symptoms are collectively called "parkinsonism", or a "parkinsonian syndrome".\\
Question: True or False: Could someone experiencing A tremor, or shaking, Slowed movement (bradykinesia), Rigid muscles, Impaired posture and balance, Loss of automatic movements, Speech changes, Writing changes. complete Volunteer for assignment and be on active duty. Have a General Technical (GT) Score of 105 or higher\\
Output:\\
Answer: False\\\\
Input:\\
Context: The Scientific Revolution was a series of events that marked the emergence of modern science during the early modern period, ...The first-generation iPhone was released on June 29, 2007, and multiple new hardware iterations with new iOS releases have been released since.\\
Question: True or False: Did 1543 occur before 2007?\\
Output:\\
Answer: False\\\\
Input:\\
Context: <<CONTEXT>>\\
Question: <<QUESTION>>\\
Output:\\
Answer: <<OUTPUT GENERATED BY GPT3>>

%% file: sections/qasc_prompt.tex
\subsection{QASC}
Prompt for original question:\\\\
Answer the given question. The question contains options A-H, choose and return the correct option. Look at the examples given below.\\\\
Input:\\
What are the vibrations in the ear called? (A) intensity (B) very complex (C) melanin content (D) lamphreys (E) Otoacoustic (F) weater (G) Seisometers (H) trucks and cars\\
Output:\\
Answer: Otoacoustic\\\\
Input:\\
<<QUESTION>>\\
Output:\\
Answer: <<OUTPUT GENERATED BY GPT3>>\\\\\\
Prompt for decomposed question:\\\\
Given a yes or no question, return yes if the answer is yes. Otherwise return no.\\
<<QUESTION>>\\
Answer: <<OUTPUT GENERATED BY GPT3>>
\input{tables/strategy_qasc}

%% file: tables/strategy_qasc.tex
\begin{table*}[htp]
    \centering 
    \setlength\tabcolsep{4.0pt}
    \footnotesize
\resizebox{0.8\linewidth}{!}{
    \begin{tabular}{cp{30em}}
    \toprule
    \multirow{9}{*}{StrategyQA}& \textbf{\emph{Context}}: Mail carriers, also referred to as mailmen or letter carriers,  …  Clothing also provides protection from ultraviolet radiation.\\
    & \textbf{\emph{Original Question}}: True or False: Mail carriers need multiple uniforms.\\
    & \textbf{\emph{Original Answer}}: True\\
    & \textbf{\emph{Generated Answer}}: False\\
    & \textbf{\emph{Decomposed Question 1}}: What seasons do mail carriers work through?\\
    & \textbf{\emph{Generated Answer}}: All seasons\\
    & \textbf{\emph{Decomposed Question 2}}: True or False: In order to make it through all of \#1, one needs multiple clothing pieces.\\
    & \textbf{\emph{Generated Answer}}: True\\
    \hline
    \multirow{7}{*}{QASC}&\\
    & \textbf{\emph{Original Question}}: what kind of beads are formed from vapor condensing? (A) h2o (B) H20 (C) tiny (D) carbon (E) hydrogen (F) rain (G) oxygen (H) Dew\\
    & \textbf{\emph{Answer}}: h2o\\
    & \textbf{\emph{Decomposed Question 1}}: Are \#1 beads formed from vapor condensing?\\
    & \textbf{\emph{Answer}}: yes\\
    \bottomrule
    \end{tabular}
    }
    \caption{Examples of decompositions for StrategyQA and QASC datasets. For each option in QASC, \#1 is replaced with the option and posed to GPT-3 as a yes or no question.}
    \label{tab:decomp3}
\end{table*}

%% file: sections/multirc_prompt.tex
\subsection{MultiRC}
Given a context-question pair, answer the question using information and facts present in the context. Keep your answers as short as possible.\\\\
Example:\\
Input:\\
Context: Should places at the same distance from the equator have the same climate? You might think they should. Unfortunately, you would not be correct to think this. Climate types vary due to other factors besides distance from the equator. So what are these factors? How can they have such a large impact on local climates? For one thing, these factors are big. You may wonder, are they as big as a car. Think bigger. Are they bigger than a house? Think bigger. Are they bigger than a football stadium? You are still not close. We are talking about mountains and oceans. They are big features and big factors. Oceans and mountains play a huge role in climates around the world. You can see this in Figure above.  Only one of those factors is latitude, or distance from the equator.\\
Question: Name at least one factor of climate\\
Output:\\
Answer: Oceans\\\\
Example:\\
Input:\\
Context: Earth processes have not changed over time. The way things happen now is the same way things happened in the past. Mountains grow and mountains slowly wear away. The same process is at work the same as it was billions of years ago. As the environment changes, living creatures adapt. They change over time. Some organisms may not be able to adapt. They become extinct. Becoming extinct means they die out completely. Some geologists study the history of the Earth. They want to learn about Earths past. They use clues from rocks and fossils. They use these clues to make sense of events. The goal is to place things in the order they happened. They also want to know how long it took for those events to happen.\\
Question: What is one example of how the earth's processes are the same today as in the past?\\
Output:\\
Answer: Things develop and then wither away\\\\
Input:\\
Context:: <<CONTEXT>>\\
Question: <<QUESTION>>\\
Output:\\
Answer: <ANSWER GENERATED BY GPT3>>
\input{tables/multirc}

%% file: tables/multirc.tex
\begin{table*}[htp]
    \centering 
    \setlength\tabcolsep{4.0pt}
    \footnotesize

\resizebox{0.8\linewidth}{!}{
    \begin{tabular}{c p{30em}}
    \toprule
    \multirow{17}{*}{MultiRC}& \textbf{\emph{Context}}: Sometimes a full Moon moves through Earths shadow. ... The Moon glows with a dull red coloring during a total lunar eclipse.\\
    & \textbf{\emph{Original Question}}: Is it more common for the Moon to travel completely in the Earth's umbra or only partially?\\
    & \textbf{\emph{List of correct answers}}: Partially, A total eclipse is less common than partial so it is more common for the moon to travel partially in Earth's umbra\\
    & \textbf{\emph{Decomposed Question 1}}:When does the Moon travel's completely in Earth's umbra?\\
    & \textbf{\emph{Answer}}: total lunar eclipse\\
    & \textbf{\emph{Decomposed Question 2}}:When does the Moon travel's partially in Earth's umbra?\\
    & \textbf{\emph{Answer}}: partial lunar eclipse\\
    & \textbf{\emph{Decomposed Question 3}}: Which is more common \#1 or \#2?\\
    & \textbf{\emph{Answer}}: partial lunar eclipse\\
    & \textbf{\emph{Decomposed Question 4}}: Does the Moon travel partially or completely in \#3?\\
    & \textbf{\emph{Answer}}: partially\\
    \bottomrule
    \end{tabular}
    }
    \caption{Decomposition Examples for MultiRC. MultiRC has multiple correct answer and the final correct answer which gives the best metrics for the generated answer is chosen as the correct answer corresponding to the generated answer. }
    \label{tab:decomp4}
\end{table*}

%% file: sections/error_examples.tex
\section{Error Examples}
\label{app:error_examples}
This section discusses the errors generated by using decompositions. We observe two types of errors while answering decomposed questions. The final answer is wrong because previous sub-questions were answered incorrectly either because such a question has multiple correct answer, or simply because the model could not understand the question correctly.

\textbf{Context}: ... Roger David Casement (1 September 1864 - 3 August 1916), formerly known as Sir Roger Casement .... In collaboration with Roger Casement, Morel led a campaign against slavery in the Congo Free State, founding the Congo Reform Association .... The association was founded in March, 1904, by Dr. Henry Grattan Guinness (1861-1915), Edmund Dene Morel, and Roger Casement ...\\
\textbf{Question}:\\
When was the date of birth of one of the founder of Congo Reform Association?\\
\textbf{True Answer}: 1 September 1864\\
\textbf{Generated Answer}: 18 October 1914\\
\\
\textbf{Decomposed Question 1}:\\
Who is the founder of the Congo Reform Association?\\
\textbf{True Answer}: Roger Casement\\
\textbf{Generated Answer}: Henry Grattan Guinness\\
\\
\textbf{Decomposed Question 2}: When was \#1 born?\\
\textbf{True Answer}: 1 September 1864\\
\textbf{Generated Answer}: 1861\\
\\
Above is an example from HotpotQA. As can be seen from the context, Congo Reform Association had multiple founders. GPT3 did give a correct answer among a set of correct answers whereas the ground truth answer provided by the dataset was some other correct option. 

% since HotpotQA requires reasoning ability at the sentence level. 
% \cite{luo2021simple} 
% and also strength of LMs \cite{luo2022choose}. 

Below is an example of incorrect retrieval. The answer generated for the first decomposed question incorrectly returns cities taken by Ottomans as well instead of just the Venetians. Hence, the final decomposed questions returns the incorrect count.\\
\\
\textbf{Context}: In the Morean War, the Republic of Venice besieged Sinj in October 1684 and then again March and April 1685, but both times without success. In the 1685 attempt, the Venetian armies were aided by the local militia of the Republic of Poljica, who thereby rebelled against their nominal Ottoman suzerainty that had existed since 1513. In an effort to retaliate to Poljica, in June 1685, the Ottomans attacked Zadvarje, and in July 1686 Dolac and Srijane, but were pushed back, and suffered major casualties. With the help of the local population of Poljica as well as the Morlachs, the fortress of Sinj finally fell to the Venetian army on 30 September 1686. On 1 September 1687 the siege of Herceg Novi started, and ended with a Venetian victory on 30 September. Knin was taken after a twelve-day siege on 11 September 1688. The capture of the Knin Fortress marked the end of the successful Venetian campaign to expand their territory in inland Dalmatia, and it also determined much of the final border between Dalmatia and Bosnia and Herzegovina that stands today. The Ottomans would besiege Sinj again in the Second Morean War, but would be repelled. On 26 November 1690, Venice took Vrgorac, which opened the route towards Imotski and Mostar. In 1694 they managed to take areas north of the Republic of Ragusa, namely Čitluk, Gabela, Zažablje, Trebinje, Popovo, Klobuk and Metković. In the final peace treaty, Venice did relinquish the areas of Popovo polje as well as Klek and Sutorina, to maintain the pre-existing demarcation near Ragusa.\\
\textbf{Question}:\\
How many cities did Venice try to take?\\
\textbf{True Answer}: 10\\
\textbf{Generated Answer}: 3\\
\\
\textbf{Decomposed Question 1}:\\
Which cities did Venice try to take?\\
\textbf{True Answer}: Sinj, Knin, Vrgorac, Čitluk, Gabela, Zažablje, Trebinje, Popovo, Klobuk and Metković\\
\textbf{Generated Answer}: Sinj, Zadvarje, Dolac, Srijane, Knin, Vrgorac, Čitluk, Gabela, Zažablje, Trebinje, Popovo, Klobuk and Metković\\
\\
\textbf{Decomposed Question 2}: What is the count of the cities mentioned in \#1?\\
\textbf{True Answer}: 10\\
\textbf{Generated Answer}: 14\\
\\
The samples for QASC are provided without context. Without the context, the answers to some of the decomposed questions can be open ended. Certain options can be unambiguously wrong and some are unambiguously correct. Below is an example:\\
\textbf{Question}: What can knowledge of the stars be used for? (A) travel (B) art (C) as a base (D) safety (E) story telling (F) light source (G) vision (H) life\\
\textbf{True Answer}: travel\\
\textbf{Generated Answer}: art\\
\\
\textbf{Decomposed Question}: Can the knowledge of stars be used for the following: \#?\\
\\
The decomposed question for each option is posed as a yes or no question to GPT3. It returns yes for art and story telling but not for travel.

%% file: sections/automated_examples.tex
\section{Examples, Results and Details for Automation}
\label{app:automated_examples}

We attempt to automate the process of decomposition using GPT3. We use the examples from manual decomposition in the prompts given to GPT3, some of which are presented below. The results obtained from the experiments are presented in Table \ref{tab: automated_gpt3}. The generated decompositions are answered using RoBERTa-base finetuned on \squad{} dataset.\\
\input{tables/automated_GPT3}\\
In this section, we present the prompts we used while attempting to automatically generate decomposed questions using GPT3.\\
\\
The prompt for generating decompositions for DROP was as follows:\\\\
Decompose a given question by breaking it into simpler sub-questions. The answer to each subsequent sub-question should lead towards the answer of the given question. To do so, use the context provided and look at the examples.
Here are some helpful instructions:

\begin{enumerate}
    \item If the given question compares two things, best strategy is to generate sub-questions that finds the answer to each of those things and compare them in the last sub-question.
    \item Some sub-questions must contain phrases like "answer of sub-question 1".
    \item If a sub-question is an arithmetic operation, then the sub-question should be framed as {operation} ! "answer of sub-question 1" ! "answer of sub-question 2".
    \item The operation used in 3) is always one of the following: summation, difference, greater, lesser.
\end{enumerate}
Example 1:\\
Context: Mehmed built a fleet to besiege the city from the sea .Contemporary estimates of the strength of the Ottoman fleet span between about 110 ships , 145 , 160 , 200-250  to 430 . A more realistic modern estimate predicts a fleet strength of 126 ships comprising 6 large galleys, 10 ordinary galleys, 15 smaller galleys, 75 large rowing boats, and 20 horse-transports.:44 Before the siege of Constantinople, it was known that the Ottomans had the ability to cast medium-sized cannons, but the range of some pieces they were able to field far surpassed the defenders' expectations. Instrumental to this Ottoman advancement in arms production was a somewhat mysterious figure by the name of Orban , a Hungarian. One cannon designed by Orban was named \"Basilica\" and was 27 feet  long, and able to hurl a 600 lb  stone ball over a mile .\\
Question: How many ordinary galleys and large rowing boats is estimated from the fleet strength?\\
Sub-question 1: How many ordinary galleys were there?\\
Sub-question 2: How many large rowing boats were there?"\\
Sub-question 3: summation ! "answer of sub-question 1" ! "answer of sub-question 2"\\\\
Example 2:\\
Context: As of the census of 2000, there were 14,702 people, 5,771 households, and 4,097 families residing in the county.  The population density was 29 people per square mile (11/km²).  There were 7,374 housing units at an average density of 14 per square mile (6/km²).  The racial makeup of the county was 98.02\% Race (United States Census), 0.69\% Race (United States Census) or Race (United States Census), 0.35\% Race (United States Census), 0.11\% Race (United States Census), 0.05\% Race (United States Census), 0.08\% from Race (United States Census), and 0.71\% from two or more races.  0.44\% of the population were Race (United States Census) or Race (United States Census) of any race.\\
Question: How many more people than households are reported according to the census?\\
Sub-question 1: As of the 2000 census, how many people are residing in the country?\\
Sub-question 2: As of the 2000 census, how many households are reported?\\
Sub-question 3: difference !"answer of sub-question 1" ! "answer of sub-question 2"\\\\
Example 3:
Context: As of the census of 2000, there were 49,129 people, 18,878 households, and 13,629 families residing in the county.  The population density was 88 people per square mile (34/km2).  There were 21,779 housing units at an average density of 39 per square mile (15/km2).  The racial makeup of the county was 74.4\% Race (United States Census), 20.4\% Race (United States Census) or Race (United States Census), 0.60\% Race (United States Census), 1.1\% Race (United States Census), 0.15\% Race (United States Census), 1.3\% from Race (United States Census), and 2.2\% from two or more races.  3.4\% of the population were Race (United States Census) or Race (United States Census) of any race. 2.85\% of the population reported speaking Spanish language at home, while 1.51\% speak German language.
Question: How many more people are there than families?
Sub-question 1: How many people are there in the 2000 census?
Sub-question 2: How many families are recorded in the 200 census?
Sub-question 3: difference ! "answer of sub-question 1" ! "answer of sub-question 2"\\\\
Context: <<CONTEXT>>\\
Question: <<QUESTION>>\\\\
<<OUTPUT GENERATED BY GPT3>>\\\\\\
The prompt for HotpotQA was similar, except replacing the examples with instances from HotpotQA. For SVAMP, since the context was much smaller, we could give more examples. The prompt for SVAMP is as shown below:\\\\
Decompose a given question by breaking it into simpler sub-questions. The answer to each subsequent sub-question should lead towards the answer of the given question. To do so, use the context provided and look at the examples.\\\\
Here are some helpful instructions:\\
\begin{enumerate}
    \item If the given question compares two things, best strategy is to generate sub-questions that finds the answer to each of those things and compare them in the last sub-question,\\
2) Some sub-questions must contain phrases like "answer of sub-question 1".
    \item Some sub-questions must contain phrases like "answer of sub-question 1".
    \item If a sub-question is an arithmetic operation, then the sub-question should be framed as {operation} ! "answer of sub-question 1" ! "answer of sub-question 2".
    \item The operation used in 3) is always one of the following: summation, difference, greater, lesser
\end{enumerate}
Example 1:\\
Context: Jessica had 8.0 quarters in her bank . Her sister borrowed 3.0 of her quarters. How many quarters does Jessica have now?\\
Sub-question 1: How many quarters did Jessica have in her bank initially?\\
Sub-question 2: How many quarters did Jessica's sister borrow?\\
Sub-question 3: difference ! "answer of sub-question  1" ! "answer of sub-question  2"\\\\
Example 2:\\
Context: Shawn has 13.0 blocks. Mildred has with 2.0 blocks. Mildred finds another 84.0. How many blocks does Mildred end with?\\
Sub-question 1: How many blocks does Mildred start with?\\
Sub-question 2: How many blocks does Mildred find?\\
Sub-question 3: summation ! "answer of sub-question  1" ! "answer of sub-question  2"\\\\
Example 3:\\
Context: Dave was helping the cafeteria workers pick up lunch trays, but he could only carry 9.0 trays at a time. If he had to pick up 17.0 trays from one table and 55.0 trays from another. how many trips will he make?\\
Sub-question 1: How many trays did Dave have to pick up from the first table?\\
Sub-question 2: How many trays did Dave have to pick up from the second table?\\
Sub-question 3: summation ! "answer of sub-question  1" ! "answer of sub-question  2"\\
Sub-question 4: How many lunch trays could Dave carry at a time?\\
Sub-question 5: division ! "answer of sub-question  3" ! "answer of sub-question  4"\\\\
Example 4:\\
Context: Paco had 93.0 cookies. Paco ate 15.0 of them. How many cookies did Paco have left?\\
Sub-question 1: How many cookies did Paco start with?\\
Sub-question 2: How many cookies did Paco eat?\\
Sub-question 3: difference ! "answer of sub-question  1" ! "answer of sub-question  2"\\\\
Example 5:\\
Context: 43 children were riding on the bus. At the bus stop some children got off the bus. Then there were 21 children left on the bus. How many children got off the bus at the bus stop?\\
Sub-question 1: How many children were on the bus at the beginning?\\
Sub-question 2: How many children were left on the bus?\\
Sub-question 3: difference ! "answer of sub-question  1" ! "answer of sub-question  2"\\\\
Example 6:\\
Context: 28 children were riding on the bus. At the bus stop 82 children got on the bus while some got off the bus. Then there were 30 children altogether on the bus. How many more children got on the bus than those that got off?\\
Sub-question 1: How many children were on the bus at the beginning?\\
Sub-question 2: How many children were left on the bus?\\
Sub-question 3: difference ! "answer of sub-question  1" ! "answer of sub-question  2"\\\\
Example 7:\\
Context: They decided to hold the party in their backyard. If they have 11 sets of tables and each set has 13 chairs, how many chairs do they have in the backyard?\\
Sub-question 1: How many tables are there in the backyard?\\
Sub-question 2: How many chairs are on each table?\\
Sub-question 3: multiplication ! "answer of sub-question  1" ! "answer of sub-question  2"\\\\
Context: <<CONTEXT + QUESTION>>\\
\\
The examples of decompositions generated for HotpotQA, DROP and SVAMP are shown in Table \ref{tab:automated_examples}

%% file: tables/automated_GPT3.tex
\begin{table*}[h]
    \centering 
    \setlength\tabcolsep{4.0pt}
    \footnotesize
\renewcommand{\arraystretch}{1.1}{
\resizebox{0.9\linewidth}{!}{ 
\begin{tabular}{ccc|cc|cc}
\toprule
Dataset  & \multicolumn{2}{c}{F1} & \multicolumn{2}{c}{EM} & \multicolumn{2}{c}{Rouge-L}\\
\cmidrule{2-7}
%\hline
 & Baseline & Decompose & Baseline & Decompose & Baseline & Decompose\\
 \hline
 HotpotQA & \textbf{32.68} & 14.12 & \textbf{29.50} & 11.47 & \textbf{33.29} & 14.00\\
 %\hline
 DROP & \textbf{22.8} & 3.77 & \textbf{21.69} & 3.77 & \textbf{23.4} & 3.76\\
 %\hline
 SVAMP & 7.4 & \textbf{17.35} & 7.4 & \textbf{17.35} & 7.4 & \textbf{17.35}\\
 \hline
 Average & \textbf{20.96} & 11.74 & \textbf{19.53} & 10.86 & \textbf{21.36} & 11.70\\
 \bottomrule
\end{tabular}
}
}

\caption{Results obtained by using decomposed questions generated using GPT3}
\label{tab: automated_gpt3}
\end{table*}

%% file: tables/automated_examples.tex
\begin{table*}[h]
    \centering 
    \setlength\tabcolsep{4.0pt}
    \footnotesize
    
\resizebox{0.8\linewidth}{!}{
    \begin{tabular}{c p{30em}}
    \hline
    \multirow{18}{*}{DROP}& \textbf{\emph{Context}}: Hoping to rebound from their loss to the Patriots, the Raiders stayed at home for a Week 16 duel with the Houston Texans.  ... The Texans tried to rally in the fourth quarter as Brown nailed a 40-yard field goal, yet the Raiders' defense would shut down any possible attempt.\\
    & \textbf{\emph{Original Question}}: How many yards longer was the longest passing touchdown than the shortest?\\\\
    & \textbf{\emph{Decomposed Question 1}}: What was the length of the shortest touchdown pass?\\
    & \textbf{\emph{Decomposed Question 2}}:What was the length of the longest touchdown pass?\\
    & \textbf{\emph{Decomposed Question 3}}: greater  ! \#1 ! \#2\\
    \hline
    \multirow{18}{*}{DROP} & \textbf{\emph{Context}}: In 1085, Guadalajara was retaken by the Christian forces of Alfonso VI . The chronicles say that the Christian army was led by Alvar Fanez de Minaya, one of the lieutenants  of El Cid. From 1085 until the Battle of Las Navas de Tolosa in 1212, the city suffered wars against the Almoravid and the Almohad Empires. In spite of the wars, the Christian population could definitely settle down in the area thanks to the repopulation with people from the North  who received their first fuero in 1133 from Alfonso VII.In 1219, the king Fernando III gave a new fuero to the city .During the reign of Alfonso X of Castile, the protection of the king allowed the city to develop its economy by protecting merchants and allowing markets.\\
    & \textbf{\emph{Original Question}}: When did the first battle against Guadalajara take place?\\\\
    & \textbf{\emph{Decomposed Question 1}}: When was Guadalajara retaken by the Christian forces?\\
    & \textbf{\emph{Decomposed Question 2}}:Who led the Christian army?\\
    & \textbf{\emph{Decomposed Question 3}}: \#1 ! \#2\\
    \hline
    \end{tabular}
    }
    \caption{Decompositions for DROP generated using GPT3}
    \label{tab:automated_examples}
\end{table*}

%% file: sections/results.tex
\section{Results}
\label{app:results}

We tabulate the results we get for all the datasets for baseline and our proposed mechanism.

\input{tables/gpt3_rc}

\input{tables/gpt3_math}

\input{tables/gpt3_strategy_qasc}

\input{tables/roberta_squad}

\input{tables/svamp_strategy_roberta}

%% file: tables/gpt3_rc.tex
\begin{table*}[htp]
    \centering 
    \setlength\tabcolsep{4.0pt}
    \footnotesize
\renewcommand{\arraystretch}{1.1}{
\resizebox{0.9\linewidth}{!}{ 
    \begin{tabular}{ccc|cc|cc}
\toprule
Dataset & \multicolumn{2}{c}{F1} & \multicolumn{2}{c}{EM} & \multicolumn{2}{c}{Rouge-L} \\
\cmidrule{2-7}
 & Baseline & Decompose & Baseline & Decompose & Baseline & Decompose\\
 \hline
 HotpotQA  & 71.97 & \textbf{78.53} & 70 & \textbf{76} & 73.33 & \textbf{79.93}\\
 %\hline
 DROP  & 52.97 & \textbf{78.16} & 46.87 & \textbf{75.86} & 46.72 & \textbf{77.66}\\
 %\hline
 MultiRC  & 64.39 & \textbf{80.74} & 33.33 & \textbf{55.55} & 61.24 & \textbf{77.31}\\
 %\hline
 BREAK  & 66.81 & \textbf{84.54} & 58 & \textbf{74} & 62.30 & \textbf{78.56}\\
 \hline
 Average & 60.10 & \textbf{81.97} & 52.64 & \textbf{76.26} & 59.35 & \textbf{81.10}\\
 \bottomrule
\end{tabular}
}
}
\caption{Comparison of metrics for reading comprehension datasets between GPT3 baseline and Decompose\_GPT3}
    \label{tab:gpt3_rc}
\end{table*}

%% file: tables/gpt3_math.tex
\begin{table*}[htp]
    \centering 
    \setlength\tabcolsep{4.0pt}
    \footnotesize
\renewcommand{\arraystretch}{1.1}{
\resizebox{0.9\linewidth}{!}{ 
    \begin{tabular}{ccc|cc|cc}
        \toprule
Dataset  & \multicolumn{2}{c}{F1} & \multicolumn{2}{c}{EM} & \multicolumn{2}{c}{Rouge-L}\\
\cmidrule{2-7}
 & Baseline & Decompose & Baseline & Decompose & Baseline & Decompose\\
 \hline
 MATH & 31.1 & \textbf{82.5} & 27.44 & \textbf{82.22} & 23.4 & \textbf{80.85}\\
 %\hline
 SVAMP & 61.80 & \textbf{78.75} & 58.88 & \textbf{77.5} & 55 & \textbf{77.5}\\
 \hline
 Average & 46.45 & \textbf{80.62} & 43.16 & \textbf{79.86} & 39.2 & \textbf{79.17}\\
 \bottomrule
\end{tabular}
}
}
\caption{Comparison of metrics for mathematical reasoning datasets between GPT3 baseline and Decompose\_GPT3}
    \label{tab:gpt3_math}
\end{table*}

%% file: tables/gpt3_strategy_qasc.tex
\begin{table*}[htp]
    \centering 
    \setlength\tabcolsep{4.0pt}
    \footnotesize
\renewcommand{\arraystretch}{1.1}{
\resizebox{0.9\linewidth}{!}{ 
    \begin{tabular}{ccc|cc|cc}
        \toprule
Dataset  & \multicolumn{2}{c}{F1} & \multicolumn{2}{c}{EM} & \multicolumn{2}{c}{Rouge-L}\\
\cmidrule{2-7}
 & Baseline & Decompose & Baseline & Decompose & Baseline & Decompose\\
 \hline
 StrategyQA & 63.15 & \textbf{84.61} & 63.15 & \textbf{84.61} & 63.15 & \textbf{84.61}\\
 %\hline
 QASC & 75.23 & \textbf{89.52} & 75.23 & \textbf{89.52} & 71.4 & \textbf{85.71}\\
 \hline
 Average & 69.19 & \textbf{87.06} & 69.19 & \textbf{87.06} & 67.27 & \textbf{85.16}\\
 \bottomrule
\end{tabular}
}
}
\caption{Comparison of metrics for StrategyQA (strategic reasoning) and QASC (fact-based multichoice) between GPT3 baseline and Decompose\_GPT3}
    \label{tab:gpt3_strategy_qasc}
\end{table*}

%% file: tables/roberta_squad.tex
\begin{table*}[htp]
    \centering 
    \setlength\tabcolsep{4.0pt}
    \footnotesize
\renewcommand{\arraystretch}{1.1}{
\resizebox{0.9\linewidth}{!}{ 
    \begin{tabular}{ccc|cc|cc}
        \toprule
Dataset  & \multicolumn{2}{c|}{F1} & \multicolumn{2}{c|}{EM} & \multicolumn{2}{c}{Rouge-L}\\
 \cmidrule{2-7}
 & Baseline & Decompose & Baseline & Decompose & Baseline & Decompose\\
 \hline
 HotpotQA & 32.14 & \textbf{49.50} & 26 & \textbf{42} & 33.33 & \textbf{50.72} \\
 %\hline
 DROP & 25.56 & \textbf{66.14} & 25 & \textbf{62.5} & 25.56 & \textbf{66.14}\\
 %\hline
 MultiRC & 45.74 & \textbf{48.1} & 24.44 & \textbf{28.88} & 44.83 & \textbf{46.95}\\
 %\hline
 BREAK & 24.6 & \textbf{36.17} & 18 & \textbf{28} & 24.31 & \textbf{35.5}\\
 \hline
 Average & 23.68 & \textbf{47.65} & 20.26 & \textbf{43.96} & 28.76 & \textbf{50.74}\\
 \bottomrule
\end{tabular}
}
}

\caption{Comparison of metrics for reading comprehension datasets between baseline and decompose settings using RoBERTa-base finetuned on SQuAD.}
\label{tab:roberta_squad_1}

\end{table*}

%% file: tables/svamp_strategy_roberta.tex
\begin{table*}[htp]
    \centering 
    \setlength\tabcolsep{4.0pt}
    \footnotesize
\renewcommand{\arraystretch}{1.1}{
\resizebox{0.9\linewidth}{!}{ 
    \begin{tabular}{ccc|cc|cc}
        \toprule
Dataset  & \multicolumn{2}{c}{F1} & \multicolumn{2}{c}{EM} & \multicolumn{2}{c}{Rouge-L}\\
\cmidrule{2-7}
 & Baseline & Decompose & Baseline & Decompose & Baseline & Decompose\\
 \hline
 StrategyQA & 47.36 & \textbf{55.26} & 47.36 & \textbf{55.26} & 47.36 & \textbf{55.26} \\
 %\hline
 SVAMP & 2 & \textbf{58} & 2 & \textbf{58} & 2 & \textbf{58}\\
 \hline
 Average & 24.68 & \textbf{56.63} & 24.68 & \textbf{56.63} & 24.68 & \textbf{56.63}\\
 \bottomrule
\end{tabular}
}
}

\caption{Comparison of metrics for StrategyQA and SVAMP between baseline and decompose settings using RoBERTa-base finetuned on SQuAD. For StrategyQA, RoBERTa-base SQuAD is used to answer intermediate decompositions whereas RoBERTa-base finetuned on BoolQ is used to answer the original question and the final decomposed question}
\label{tab:roberta_squad_2}

\end{table*}

%% file: emnlp2022.bbl
\begin{thebibliography}{36}
\expandafter\ifx\csname natexlab\endcsname\relax\def\natexlab#1{#1}\fi

\bibitem[{Amini et~al.(2019)Amini, Gabriel, Lin, Koncel-Kedziorski, Choi, and
  Hajishirzi}]{mathqa}
Aida Amini, Saadia Gabriel, Shanchuan Lin, Rik Koncel-Kedziorski, Yejin Choi,
  and Hannaneh Hajishirzi. 2019.
\newblock Mathqa: Towards interpretable math word problem solving with
  operation-based formalisms.
\newblock \emph{ArXiv}, abs/1905.13319.

\bibitem[{Brown et~al.(2020)Brown, Mann, Ryder, Subbiah, Kaplan, Dhariwal,
  Neelakantan, Shyam, Sastry, Askell, Agarwal, Herbert{-}Voss, Krueger,
  Henighan, Child, Ramesh, Ziegler, Wu, Winter, Hesse, Chen, Sigler, Litwin,
  Gray, Chess, Clark, Berner, McCandlish, Radford, Sutskever, and
  Amodei}]{gpt3}
Tom~B. Brown, Benjamin Mann, Nick Ryder, Melanie Subbiah, Jared Kaplan,
  Prafulla Dhariwal, Arvind Neelakantan, Pranav Shyam, Girish Sastry, Amanda
  Askell, Sandhini Agarwal, Ariel Herbert{-}Voss, Gretchen Krueger, Tom
  Henighan, Rewon Child, Aditya Ramesh, Daniel~M. Ziegler, Jeffrey Wu, Clemens
  Winter, Christopher Hesse, Mark Chen, Eric Sigler, Mateusz Litwin, Scott
  Gray, Benjamin Chess, Jack Clark, Christopher Berner, Sam McCandlish, Alec
  Radford, Ilya Sutskever, and Dario Amodei. 2020.
\newblock \href {http://arxiv.org/abs/2005.14165} {Language models are few-shot
  learners}.
\newblock \emph{CoRR}, abs/2005.14165.

\bibitem[{Clark et~al.(2019)Clark, Lee, Chang, Kwiatkowski, Collins, and
  Toutanova}]{clark2019boolq}
Christopher Clark, Kenton Lee, Ming-Wei Chang, Tom Kwiatkowski, Michael
  Collins, and Kristina Toutanova. 2019.
\newblock Boolq: Exploring the surprising difficulty of natural yes/no
  questions.
\newblock \emph{arXiv preprint arXiv:1905.10044}.

\bibitem[{Dua et~al.(2019)Dua, Dasigi, Stanovsky, Singh, and
  Gardner}]{DROP:journals/corr/abs-1903-00161}
Dheeru Dua, Yizhong Wang~Pradeep Dasigi, Gabriel Stanovsky, Sameer Singh, and
  Matt Gardner. 2019.
\newblock \href {http://arxiv.org/abs/1903.00161} {{DROP:} {A} reading
  comprehension benchmark requiring discrete reasoning over paragraphs}.
\newblock \emph{CoRR}, abs/1903.00161.

\bibitem[{Geva et~al.(2021)Geva, Khashabi, Segal, Khot, Roth, and
  Berant}]{strategyqa}
Mor Geva, Daniel Khashabi, Elad Segal, Tushar Khot, Dan Roth, and Jonathan
  Berant. 2021.
\newblock Did aristotle use a laptop? a question answering benchmark with
  implicit reasoning strategies.
\newblock \emph{Transactions of the Association for Computational Linguistics},
  9:346--361.

\bibitem[{Guo et~al.(2019)Guo, Zhan, Gao, Xiao, Lou, Liu, and Zhang}]{text2sql}
Jiaqi Guo, Zecheng Zhan, Yan Gao, Yan Xiao, Jian-Guang Lou, Ting Liu, and
  Dongmei Zhang. 2019.
\newblock Towards complex text-to-sql in cross-domain database with
  intermediate representation.
\newblock \emph{arXiv preprint arXiv:1905.08205}.

\bibitem[{Gupta and Lewis(2018)}]{denotational_semantics_gupta}
Nitish Gupta and Mike Lewis. 2018.
\newblock Neural compositional denotational semantics for question answering.
\newblock \emph{arXiv preprint arXiv:1808.09942}.

\bibitem[{Iyyer et~al.(2017)Iyyer, Yih, and Chang}]{iyyer-search-based}
Mohit Iyyer, Wen-tau Yih, and Ming-Wei Chang. 2017.
\newblock \href {https://doi.org/10.18653/v1/P17-1167} {Search-based neural
  structured learning for sequential question answering}.
\newblock In \emph{Proceedings of the 55th Annual Meeting of the Association
  for Computational Linguistics (Volume 1: Long Papers)}, pages 1821--1831,
  Vancouver, Canada. Association for Computational Linguistics.

\bibitem[{Khashabi et~al.(2018)Khashabi, Chaturvedi, Roth, Upadhyay, and
  Roth}]{MultiRC2018}
Daniel Khashabi, Snigdha Chaturvedi, Michael Roth, Shyam Upadhyay, and Dan
  Roth. 2018.
\newblock Looking beyond the surface:a challenge set for reading comprehension
  over multiple sentences.
\newblock In \emph{NAACL}.

\bibitem[{Khot et~al.(2020{\natexlab{a}})Khot, Clark, Guerquin, Jansen, and
  Sabharwal}]{qasc}
Tushar Khot, Peter Clark, Michal Guerquin, Peter~Alexander Jansen, and Ashish
  Sabharwal. 2020{\natexlab{a}}.
\newblock Qasc: A dataset for question answering via sentence composition.
\newblock In \emph{AAAI}.

\bibitem[{Khot et~al.(2020{\natexlab{b}})Khot, Khashabi, Richardson, Clark, and
  Sabharwal}]{modularqa:journals/corr/abs-2009-00751}
Tushar Khot, Daniel Khashabi, Kyle Richardson, Peter Clark, and Ashish
  Sabharwal. 2020{\natexlab{b}}.
\newblock \href {http://arxiv.org/abs/2009.00751} {Text modular networks:
  Learning to decompose tasks in the language of existing models}.
\newblock \emph{CoRR}, abs/2009.00751.

\bibitem[{Kuznia et~al.(2022)Kuznia, Mishra, Parmar, and
  Baral}]{kuznia2022less}
Kirby Kuznia, Swaroop Mishra, Mihir Parmar, and Chitta Baral. 2022.
\newblock Less is more: Summary of long instructions is better for program
  synthesis.
\newblock \emph{arXiv preprint arXiv:2203.08597}.

\bibitem[{Lewis et~al.(2020)Lewis, Liu, Goyal, Ghazvininejad, Mohamed, Levy,
  Stoyanov, and Zettlemoyer}]{lewis2020bart}
Mike Lewis, Yinhan Liu, Naman Goyal, Marjan Ghazvininejad, Abdelrahman Mohamed,
  Omer Levy, Veselin Stoyanov, and Luke Zettlemoyer. 2020.
\newblock Bart: Denoising sequence-to-sequence pre-training for natural
  language generation, translation, and comprehension.
\newblock In \emph{Proceedings of the 58th Annual Meeting of the Association
  for Computational Linguistics}, pages 7871--7880.

\bibitem[{Lin(2004)}]{lin2004rouge}
Chin-Yew Lin. 2004.
\newblock Rouge: A package for automatic evaluation of summaries.
\newblock In \emph{Text summarization branches out}, pages 74--81.

\bibitem[{Liu et~al.(2021)Liu, Yuan, Fu, Jiang, Hayashi, and
  Neubig}]{liu2021pre}
Pengfei Liu, Weizhe Yuan, Jinlan Fu, Zhengbao Jiang, Hiroaki Hayashi, and
  Graham Neubig. 2021.
\newblock Pre-train, prompt, and predict: A systematic survey of prompting
  methods in natural language processing.
\newblock \emph{arXiv preprint arXiv:2107.13586}.

\bibitem[{Liu et~al.(2019)Liu, Ott, Goyal, Du, Joshi, Chen, Levy, Lewis,
  Zettlemoyer, and Stoyanov}]{liu2019roberta}
Yinhan Liu, Myle Ott, Naman Goyal, Jingfei Du, Mandar Joshi, Danqi Chen, Omer
  Levy, Mike Lewis, Luke Zettlemoyer, and Veselin Stoyanov. 2019.
\newblock Roberta: A robustly optimized bert pretraining approach.
\newblock \emph{arXiv preprint arXiv:1907.11692}.

\bibitem[{Luo et~al.(2022)Luo, Saxena, Mishra, Parmar, and
  Baral}]{luo2022biotabqa}
Man Luo, Sharad Saxena, Swaroop Mishra, Mihir Parmar, and Chitta Baral. 2022.
\newblock Biotabqa: Instruction learning for biomedical table question
  answering.
\newblock \emph{arXiv preprint arXiv:2207.02419}.

\bibitem[{Min et~al.(2019)Min, Zhong, Zettlemoyer, and Hajishirzi}]{decomprc}
Sewon Min, Victor Zhong, Luke Zettlemoyer, and Hannaneh Hajishirzi. 2019.
\newblock Multi-hop reading comprehension through question decomposition and
  rescoring.
\newblock \emph{arXiv preprint arXiv:1906.02916}.

\bibitem[{Mishra et~al.(2021{\natexlab{a}})Mishra, Khashabi, Baral, Choi, and
  Hajishirzi}]{mishra2021reframing}
Swaroop Mishra, Daniel Khashabi, Chitta Baral, Yejin Choi, and Hannaneh
  Hajishirzi. 2021{\natexlab{a}}.
\newblock Reframing instructional prompts to gptk's language.
\newblock \emph{ACL Findings}.

\bibitem[{Mishra et~al.(2021{\natexlab{b}})Mishra, Khashabi, Baral, and
  Hajishirzi}]{mishra2021cross}
Swaroop Mishra, Daniel Khashabi, Chitta Baral, and Hannaneh Hajishirzi.
  2021{\natexlab{b}}.
\newblock Cross-task generalization via natural language crowdsourcing
  instructions.
\newblock \emph{ACL}.

\bibitem[{Mishra et~al.(2022)Mishra, Mitra, Varshney, Sachdeva, Clark, Baral,
  and Kalyan}]{mishra2022numglue}
Swaroop Mishra, Arindam Mitra, Neeraj Varshney, Bhavdeep Sachdeva, Peter Clark,
  Chitta Baral, and Ashwin Kalyan. 2022.
\newblock Numglue: A suite of fundamental yet challenging mathematical
  reasoning tasks.
\newblock In \emph{Proceedings of the 60th Annual Meeting of the Association
  for Computational Linguistics (Volume 1: Long Papers)}, pages 3505--3523.

\bibitem[{Ouyang et~al.(2022)Ouyang, Wu, Jiang, Almeida, Wainwright, Mishkin,
  Zhang, Agarwal, Slama, Ray et~al.}]{ouyang2022training}
Long Ouyang, Jeff Wu, Xu~Jiang, Diogo Almeida, Carroll~L Wainwright, Pamela
  Mishkin, Chong Zhang, Sandhini Agarwal, Katarina Slama, Alex Ray, et~al.
  2022.
\newblock Training language models to follow instructions with human feedback.
\newblock \emph{arXiv preprint arXiv:2203.02155}.

\bibitem[{Parmar et~al.(2022)Parmar, Mishra, Purohit, Luo, Murad, and
  Baral}]{parmar2022boxbart}
Mihir Parmar, Swaroop Mishra, Mirali Purohit, Man Luo, M~Hassan Murad, and
  Chitta Baral. 2022.
\newblock {In-BoXBART: Get Instructions into Biomedical Multi-Task Learning}.
\newblock \emph{NAACL 2022 Findings}.

\bibitem[{Patel et~al.(2021)Patel, Bhattamishra, and Goyal}]{svamp}
Arkil Patel, Satwik Bhattamishra, and Navin Goyal. 2021.
\newblock Are nlp models really able to solve simple math word problems?
\newblock \emph{arXiv preprint arXiv:2103.07191}.

\bibitem[{Perez et~al.(2020)Perez, Lewis, Yih, Cho, and
  Kiela}]{perez2020unsupervised}
Ethan Perez, Patrick Lewis, Wen-tau Yih, Kyunghyun Cho, and Douwe Kiela. 2020.
\newblock Unsupervised question decomposition for question answering.
\newblock \emph{arXiv preprint arXiv:2002.09758}.

\bibitem[{Puri et~al.(2022)Puri, Mishra, Parmar, and Baral}]{puri2022many}
Ravsehaj~Singh Puri, Swaroop Mishra, Mihir Parmar, and Chitta Baral. 2022.
\newblock How many data samples is an additional instruction worth?
\newblock \emph{arXiv preprint arXiv:2203.09161}.

\bibitem[{Radford et~al.(2019)Radford, Wu, Child, Luan, Amodei, and
  Sutskever}]{gpt2}
Alec Radford, Jeff Wu, Rewon Child, David Luan, Dario Amodei, and Ilya
  Sutskever. 2019.
\newblock Language models are unsupervised multitask learners.

\bibitem[{Rajpurkar et~al.(2018)Rajpurkar, Jia, and Liang}]{rajpurkar2018know}
Pranav Rajpurkar, Robin Jia, and Percy Liang. 2018.
\newblock Know what you don't know: Unanswerable questions for squad.
\newblock \emph{arXiv preprint arXiv:1806.03822}.

\bibitem[{Sanh et~al.(2021{\natexlab{a}})Sanh, Webson, Raffel, Bach, Sutawika,
  Alyafeai, Chaffin, Stiegler, Scao, Raja, Dey, Bari, Xu, Thakker, Sharma,
  Szczechla, Kim, Chhablani, Nayak, Datta, Chang, Jiang, Wang, Manica, Shen,
  Yong, Pandey, Bawden, Wang, Neeraj, Rozen, Sharma, Santilli, Fevry, Fries,
  Teehan, Biderman, Gao, Bers, Wolf, and Rush}]{t0pp}
Victor Sanh, Albert Webson, Colin Raffel, Stephen~H. Bach, Lintang Sutawika,
  Zaid Alyafeai, Antoine Chaffin, Arnaud Stiegler, Teven~Le Scao, Arun Raja,
  Manan Dey, M~Saiful Bari, Canwen Xu, Urmish Thakker, Shanya~Sharma Sharma,
  Eliza Szczechla, Taewoon Kim, Gunjan Chhablani, Nihal Nayak, Debajyoti Datta,
  Jonathan Chang, Mike Tian-Jian Jiang, Han Wang, Matteo Manica, Sheng Shen,
  Zheng~Xin Yong, Harshit Pandey, Rachel Bawden, Thomas Wang, Trishala Neeraj,
  Jos Rozen, Abheesht Sharma, Andrea Santilli, Thibault Fevry, Jason~Alan
  Fries, Ryan Teehan, Stella Biderman, Leo Gao, Tali Bers, Thomas Wolf, and
  Alexander~M. Rush. 2021{\natexlab{a}}.
\newblock \href {http://arxiv.org/abs/2110.08207} {Multitask prompted training
  enables zero-shot task generalization}.

\bibitem[{Sanh et~al.(2021{\natexlab{b}})Sanh, Webson, Raffel, Bach, Sutawika,
  Alyafeai, Chaffin, Stiegler, Scao, Raja et~al.}]{sanh2021multitask}
Victor Sanh, Albert Webson, Colin Raffel, Stephen~H Bach, Lintang Sutawika,
  Zaid Alyafeai, Antoine Chaffin, Arnaud Stiegler, Teven~Le Scao, Arun Raja,
  et~al. 2021{\natexlab{b}}.
\newblock Multitask prompted training enables zero-shot task generalization.
\newblock \emph{arXiv preprint arXiv:2110.08207}.

\bibitem[{Talmor and Berant(2018)}]{complexwebqa}
Alon Talmor and Jonathan Berant. 2018.
\newblock The web as a knowledge-base for answering complex questions.
\newblock \emph{arXiv preprint arXiv:1803.06643}.

\bibitem[{Wei et~al.(2021)Wei, Bosma, Zhao, Guu, Yu, Lester, Du, Dai, and
  Le}]{wei2021finetuned}
Jason Wei, Maarten Bosma, Vincent~Y Zhao, Kelvin Guu, Adams~Wei Yu, Brian
  Lester, Nan Du, Andrew~M Dai, and Quoc~V Le. 2021.
\newblock Finetuned language models are zero-shot learners.
\newblock \emph{arXiv preprint arXiv:2109.01652}.

\bibitem[{Wei et~al.(2022)Wei, Wang, Schuurmans, Bosma, Chi, Le, and
  Zhou}]{wei2022chain}
Jason Wei, Xuezhi Wang, Dale Schuurmans, Maarten Bosma, Ed~Chi, Quoc Le, and
  Denny Zhou. 2022.
\newblock Chain of thought prompting elicits reasoning in large language
  models.
\newblock \emph{arXiv preprint arXiv:2201.11903}.

\bibitem[{Wolfson et~al.(2020)Wolfson, Geva, Gupta, Gardner, Goldberg, Deutch,
  and Berant}]{break}
Tomer Wolfson, Mor Geva, Ankit Gupta, Matt Gardner, Yoav Goldberg, Daniel
  Deutch, and Jonathan Berant. 2020.
\newblock Break it down: A question understanding benchmark.
\newblock \emph{Transactions of the Association for Computational Linguistics},
  8:183--198.

\bibitem[{Xie et~al.(2022)Xie, Wiegreffe, and Riedl}]{decompositional_probes}
Kaige Xie, Sarah Wiegreffe, and Mark Riedl. 2022.
\newblock \href {https://doi.org/10.48550/ARXIV.2204.07693} {Calibrating trust
  of multi-hop question answering systems with decompositional probes}.

\bibitem[{Yang et~al.(2018)Yang, Qi, Zhang, Bengio, Cohen, Salakhutdinov, and
  Manning}]{hotpot:journals/corr/abs-1809-09600}
Zhilin Yang, Peng Qi, Saizheng Zhang, Yoshua Bengio, William~W. Cohen, Ruslan
  Salakhutdinov, and Christopher~D. Manning. 2018.
\newblock \href {http://arxiv.org/abs/1809.09600} {Hotpotqa: {A} dataset for
  diverse, explainable multi-hop question answering}.
\newblock \emph{CoRR}, abs/1809.09600.

\end{thebibliography}
